\documentclass[10pt,twocolumn,letterpaper]{article}

\usepackage[pagenumbers]{cvpr} 
\usepackage{booktabs}
\usepackage{graphicx}
\usepackage{enumitem}
\usepackage{algorithm}
\usepackage{amsmath, amsfonts}
\usepackage{multirow}
\usepackage{ragged2e}

\usepackage[pagebackref,breaklinks,colorlinks]{hyperref}

\usepackage[capitalize]{cleveref}
\crefname{section}{Sec.}{Secs.}
\Crefname{section}{Section}{Sections}
\Crefname{table}{Table}{Tables}
\crefname{table}{Tab.}{Tabs.}

\def\cvprPaperID{*****} 
\def\confName{CVPR}
\def\confYear{2022}

\let\origtau\tau
\renewcommand{\tau}{\scalebox{1.44}{$\origtau$}}

\begin{document}

\title{Cross-Stream Contrastive Learning for Self-Supervised Skeleton-Based Action Recognition}%

\author{Ding Li\textsuperscript{1,3}, Yongqiang Tang\textsuperscript{3}, Zhizhong Zhang\textsuperscript{2}, Wensheng Zhang\textsuperscript{3}\\
\textsuperscript{1} School of Artificial Intelligence, UCAS \quad \textsuperscript{2} East China Normal University \\ \textsuperscript{3} State Key Laboratory of Multimodal Artificial Intelligence Systems, Institute of Automation, CAS
\\
{\tt\small liding2019@ia.ac.cn, yongqiang.tang@ia.ac.cn, zzzhang@cs.ecnu.edu.cn}
}
\maketitle

\begin{abstract}
   Self-supervised skeleton-based action recognition enjoys a rapid growth along with the development of contrastive learning. The existing methods rely on imposing invariance to augmentations of 3D skeleton within a single data stream, which merely leverages the easy positive pairs and limits the ability to explore the complicated movement patterns. In this paper, we advocate that the defect of single-stream contrast and the lack of necessary feature transformation are responsible for easy positives, and therefore propose a Cross-Stream Contrastive Learning framework for skeleton-based action Representation learning (CSCLR). Specifically, the proposed CSCLR not only utilizes intra-stream contrast pairs, but introduces inter-stream contrast pairs as hard samples to formulate a better representation learning. Besides, to further exploit the potential of positive pairs and increase the robustness of self-supervised representation learning, we propose a Positive Feature Transformation (PFT) strategy which adopts feature-level manipulation to increase the variance of positive pairs. To validate the effectiveness of our method, we conduct extensive experiments on three benchmark datasets NTU-RGB+D 60, NTU-RGB+D 120 and PKU-MMD. Experimental results show that our proposed CSCLR exceeds the state-of-the-art methods on a diverse range of evaluation protocols.
\end{abstract}

\section{Introduction}
\label{sec-1}
Skeleton-based action recognition has always attracted considerable research interests in the filed of computer vision, as it plays a significant role in many real-world applications, such as smart surveillance, human-machine interaction and mixed reality \cite{sun2021multi, liu2017skeleton, presti20163d, freire2022inflated}. It aims to recognize human actions using skeleton keypoints, and shows advantages under dynamic circumstance with complicated background (\emph{e.g.} clutter scene, light-conditions) \cite{yan2018spatial, shi2019two}. In the past years, most existing skeleton-based action recognition methods are based on supervised learning paradigm, which requires immense time and manual effort for annotating. Instead, self-supervised learning paradigm avoids such limitations, aiming to learn discriminative spatio-temporal action representations by exploring unlabeled skeleton data.

Several self-supervised approaches formulate the pretext task in the way of Generative Learning, including reconstruction of input skeleton sequence \cite{zheng2018unsupervised}, solving jigsaw puzzles \cite{lin2020ms2l} and motion prediction \cite{cheng2021motion}. However, the generative pretext tasks force the model to pay excessive attention to low-level detailed joint features, while ignoring the high-level semantic information which is more critical to downstream tasks. Different from the pretext task in generative methods, Contrastive Learning typically leverages the instance discrimination of skeleton sequences in the feature space. Different augmentations are applied to skeleton sequences, thus generating multiple views. Then, contrastive pairs are constructed based on these views, and the inherent consistency constraints is utilized to attract the positive pairs and repel negative pairs simultaneously \cite{he2020momentum, chen2020simple}. Based on the high-level semantic features, the contrastive learning enables the model own its advantage in downstream tasks.

\begin{figure}[t]
	\centering
    \includegraphics[width=0.48\textwidth]{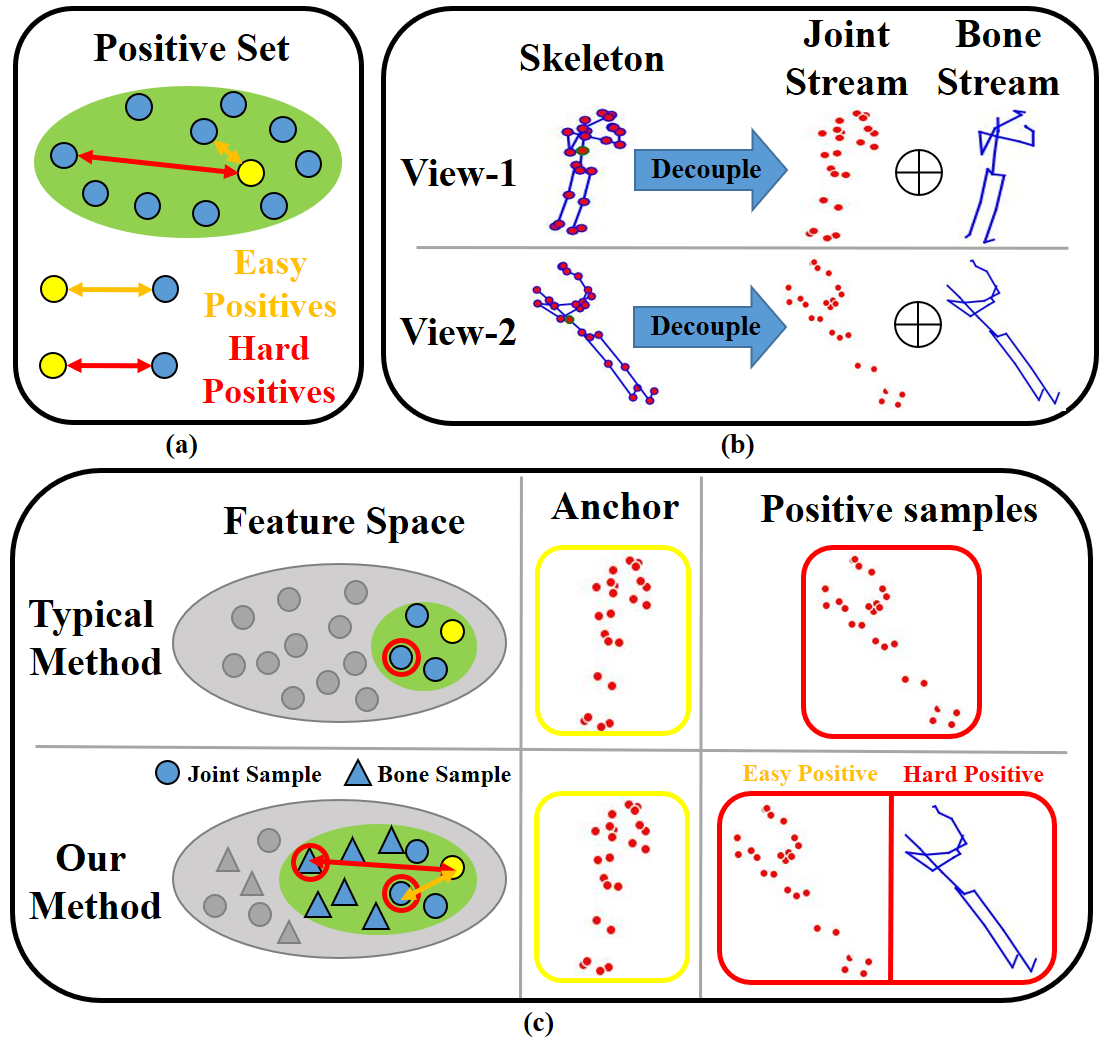}
	\caption{\textbf{An illustration of the proposed approach.} (a) Easy positive pairs distribute closely in feature space, while hard positive pairs are relatively far away from each other. (b) A skeleton sequence can be augmented to formulate two views, and then joint and bone stream are decoupled from the two views. (c) Comparing with typical method, CSCLR enforces not only intra-stream correspondence (easy positive), but correspondence between different streams (hard positive), $e.g.$ joint stream and bone stream.}
\label{fig1}
\end{figure}

Despite the success of contrastive learning in this area, previous methods rely on imposing invariance to augmentations of 3D skeleton, and these positive pairs still could  be  quite  similar in the feature space, $i.e.$, \emph{easy positives}. As shown in Figure \ref{fig1}(a), easy positives are distributed closely, which results in less and less contribution to the loss as training progresses. Using the easy positive pairs, the contrastive pretext task will be easily-accomplished, and thus limiting the ability to explore the complicated movement patterns \cite{guo2022contrastive, tian2020makes, chen2020improved}. On contrary, compared with easy positives, hard positives have longer distance in feature space and contribute more to the loss. Obviously, the hard positives, which  could make contrastive pretext task tougher in turn, are expected to promote learning more robust representation.

In this study, we mainly focus on handling two important issues that lead to high similarity of positive pairs: \emph{the defect of single-stream contrast} and \emph{the lack of necessary feature transformation}. First, among existing works, only a single skeleton data stream is used for designing contrast pairs, while other streams are underutilized. Actually, based on the raw data, multiple streams ($e.g.$ joint, bone and motion stream) can be obtained. There have been some studies dedicating to fuse the prediction results of these streams in the downstream tasks \cite{guo2022contrastive, li20213d, yang2021skeleton, shi2019two, liu2020disentangling}. However, in contrastive pre-training stage, previous methods generally overlook the multi-stream scenario and only construct pairs within a single data stream. The pairs, which are derived from the same stream, share the similar inherent information and tend to be easy positives. Second, conventional methods typically utilize data augmentations to construct positive pairs \cite{li20213d, guo2022contrastive, yang2021skeleton, rao2021augmented, thoker2021skeleton}, but such data-level operation of these pairs may not be able to explicitly guarantee their discrepancy in feature level. At the same time, multiple sample views can be designed not only based on input data space, but on feature space as well \cite{verma2018manifold, zhou2021domain, ding2021kfc}. Compared with data augmentation, feature transformation offers an explicit solution to design more effective pairs for training. Thus, by involving the feature-level manipulation, the similarity of positive pairs are expected to be further reduced. 

To remedy the issues mentioned above, we propose \textbf{CSCLR}: a cross-stream contrastive learning framework with features extracted from multiple data stream. 
For the first issue, CSCLR exploits the correspondence between different skeleton data streams, so as to constructively learn transferable skeleton representations that benefit to downstream tasks. To be specific, as shown in Figure \ref{fig1}(b), we generate bone stream accompanied with joint stream. 
And then, different from typical methods that only conduct intra-stream contrastive learning, our CSCLR additionally constructs inter-stream contrast pairs, $i.e.$, the two embeddings in each pair come from different source data stream. Since the inherent information in other data streams are different from the joint stream, thus the inter-stream contrast pairs are expected to act as hard positives to formulate a better representation learning (see  Figure \ref{fig1}(c)).
For the second issue, we propose the Positive Feature Transformation (PFT) strategy, which aims to increase the variance of features in positive set through feature extrapolation. Inspired by Mixup \cite{zhang2017mixup} and Manifold Mixup \cite{verma2018manifold}, PFT applies linear extrapolation on the paired query and key features and generate synthetic positive features for contrastive learning. After PFT, the hardness of these positive samples are enhanced, leading to evident gains in representation learning.

The main contributions of this paper can be summarized as follows:
\begin{enumerate}
\item We propose a novel cross-stream contrastive learning model named CSCLR for self-supervised skeleton-based action recognition, which contrasts the pairwise features extracted from different data stream. Compared with existing methods only using intra-stream contrast, CSCLR is proposed to introduce more hard samples from other streams, thus resulting in stronger effect of contrastive learning and better generalization performance in downstream tasks.
\item To obtain more robust skeleton feature, we design a positive feature transformation strategy, which manipulates the positive pairs to increase their hardness in feature-level. A further enhancement of contrastive learning is achieved with the generated synthetic positive features, enabling the skeleton encoder to learn more effective representations.
\item We conduct extensive experiments and ablation study on three popular benchmark datasets, $i.e.$, NTU-RGBD-60, NTU-RGBD-120 and PKU-MMD. As a result, our CSCLR achieves the state-of-the-art performance under variety of evaluation protocols, including linear evaluation, semi-supervised evaluation, and fintuned evaluation.
\end{enumerate}

The rest of this article is organized as follows. Section \ref{sec-2} introduces related works. In Section \ref{sec-3}, we introduce our method with detailed description. Experimental analysis and comparison results are shown in Section \ref{sec-4} to verify our method. Finally, we conclude this paper in Section \ref{sec-5}.


\section{Related Work}\label{sec-2}
In order to indicate our proposed method, we will review related advances in three research areas in this section, including contrastive self-supervised representation learning, supervised skeleton-based action recognition and self-supervised skeleton-based action recognition.

\subsection{Contrastive Self-Supervised Representation Learning}
Contrastive self-supervised learning aims to learn feature representations from unlabeled data with contrastive loss. DIM \cite{hjelm2018learning} maximizes the mutual information between a region of input to the encoder and its output. MoCo \cite{he2020momentum, chen2020improved} builds a dynamic dictionary with a queue and a moving-averaged encoder for computing the contrastive loss, which enables building a large and consistent dictionary on-the-fly that facilitates contrastive unsupervised learning. SimCLR \cite{chen2020simple, chen2020big} does not adopt memory bank and introduces a non-linear transformation between the representation and the loss function. PIRL \cite{misra2020self} develops pretext-Invariant method that learns invariant representations based on pretext tasks, and substantially improves the semantic quality of the learned image representations. InfoCL \cite{Wang_2022_CVPR} claims that non-shared task-relevant information cannot be ignored and proposes to increase the mutual information between the representation and input as regularization to approximately introduce more task-relevant information. CrossPoint \cite{afham2022crosspoint} constructs a joint training objective which combines the feature correspondences within and across modalities, thus ensembles a rich learning signal from both 3D point cloud and 2D image modalities in a self-supervised fashion. \cite{zhu2021improving} visualizes the similarity score distributions of pairwise samples in contrastive learning, and generates more effective contrast pairs for training. Contrastive learning with multiple inputs, $e.g.$ image + audio \cite{arandjelovic2017look, morgado2021audio, owens2018audio}, video + sentence \cite{sariyildiz2020learning, radford2021learning, desai2021virtex}, has also achieved favorable advances recently. The advances in this research area lay a solid foundation for our work, and show a promising direction for us to delve into hard contrast pairs based on different streams of 3D skeleton data.

\subsection{Supervised Skeleton-Based Action Recognition}
The input of skeleton-based action recognition is 2D/3D skeleton data, which is robust to illumination change, scene variation, and complex backgrounds \cite{yan2018spatial, shi2019two}. Traditional methods adopt hand-crafted features from raw skeleton sequences, such as \cite{wang2012mining, vemulapalli2016rolling, vemulapalli2014human}. Inspired by the success of deep learning, numerous methods based on deep neural network are carried out to extract discriminative skeleton features. Among these deep learning methods, the most widely-used networks are recurrent neural networks (RNN), convolutional neural networks (CNN) and graph convolutional networks (GCN). RNN-based models \cite{du2015hierarchical, song2017end, li2018independently} usually concatenate the coordinates of all joints in each frame as a vector and then take the sequence of vectors as input of RNN. Due to the gradient vanishing in RNN-based methods \cite{hochreiter2001gradient}, researchers would pay attention to CNN-based models. CNN-based methods \cite{ke2017new, li2017skeleton} usually transfer the raw skeleton sequence into image form, and then adopt convolution operations on this image to capture effective representation. 

Recently, GCN-based methods \cite{yan2018spatial, shi2019two, liu2020disentangling, chen2021channel, chi2022infogcn} are proposed to model the skeleton data as graph with nodes and edges, resulting in great success in this task. ST-GCN \cite{yan2018spatial} proposes a generic graph-based formulation for modeling dynamic skeletons and introduces several principles in designing convolution kernels to meet the specific demands. 2s-AGCN \cite{shi2019two} is proposed to model first-order and the second-order information simultaneously with joint and bone data stream, which shows notable improvement for the recognition accuracy. We choose to use the widely-used ST-GCN as the encoder in this paper, and the easily-obtained data streams ($e.g.$ joint, bone, motion) to facilitate the multi-stream learning greatly.

\subsection{Self-Supervised Skeleton-Based Action Recognition} Inspired by the success of self-supervised learning in image and video tasks \cite{he2020momentum, chen2020simple, pan2021videomoco, tgl9713748, huang2021self}, many self-supervised methods are proposed to capture effective skeleton representations. Previous methods can be divided into two categories by the input data: single-stream and multi-stream. 

The single-stream methods usually only use joint-stream data as input. For example, LongT GAN \cite{zheng2018unsupervised} proposes a conditional skeleton inpainting architecture for learning a fixed-dimensional representation, and utilizes the encoder-decoder framework to regenerate the skeleton sequence with a adversarial strategy. MS$^2$L \cite{lin2020ms2l} integrates multiple pretext tasks to learn more general features. P\&C \cite{su2020predict} proposes a novel training strategy which weakens the decoder and forces the encoder to learn a more informative representation. AS-CAL \cite{rao2021augmented} exploits different augmentations of unlabeled skeleton sequences to learn action representations. ISC \cite{thoker2021skeleton} integrates information between multiple forms of encoders to learn better features. ST-CL \cite{gao2021efficient} explores the pretext task with different spatio-temporal observation scenes and devises a efficient action encoder. SeBiReNet \cite{nie2020unsupervised} considers both the kinematic and geometric dependencies and design a sequential bidirectional recursive network. 

Multi-stream methods fuse the predictions of multiple data streams in downstream tasks, thus achieving better performance. SkeletonCLR \cite{li20213d} adopts the MoCo pipeline and designs a simple baseline for this task. CrosSCLR \cite{li20213d} proposes a cross-stream knowledge mining strategy to enlarge the positive set for contrastive learning, where potential positive samples are mined with the reference of feature similarity in other data streams. AimCLR \cite{guo2022contrastive} introduces the extremely-augmented skeleton sequences and proposes a nearest neighbor mining to discover the potential positive sample in the memory bank. However, both single-stream methods and multi-stream methods are dragged down by easy positives. For one thing, the positive pairs in these methods are still constructed within a single data stream, indicating the correspondence between different data streams is overlooked in contrastive learning. For another, these works lack of necessary manipulation to enhance the hardness of positive features. 

\section{Methods}\label{sec-3}
\begin{figure*}[!t]
	\centering
    \includegraphics[width=17.5cm,height=8.3cm]{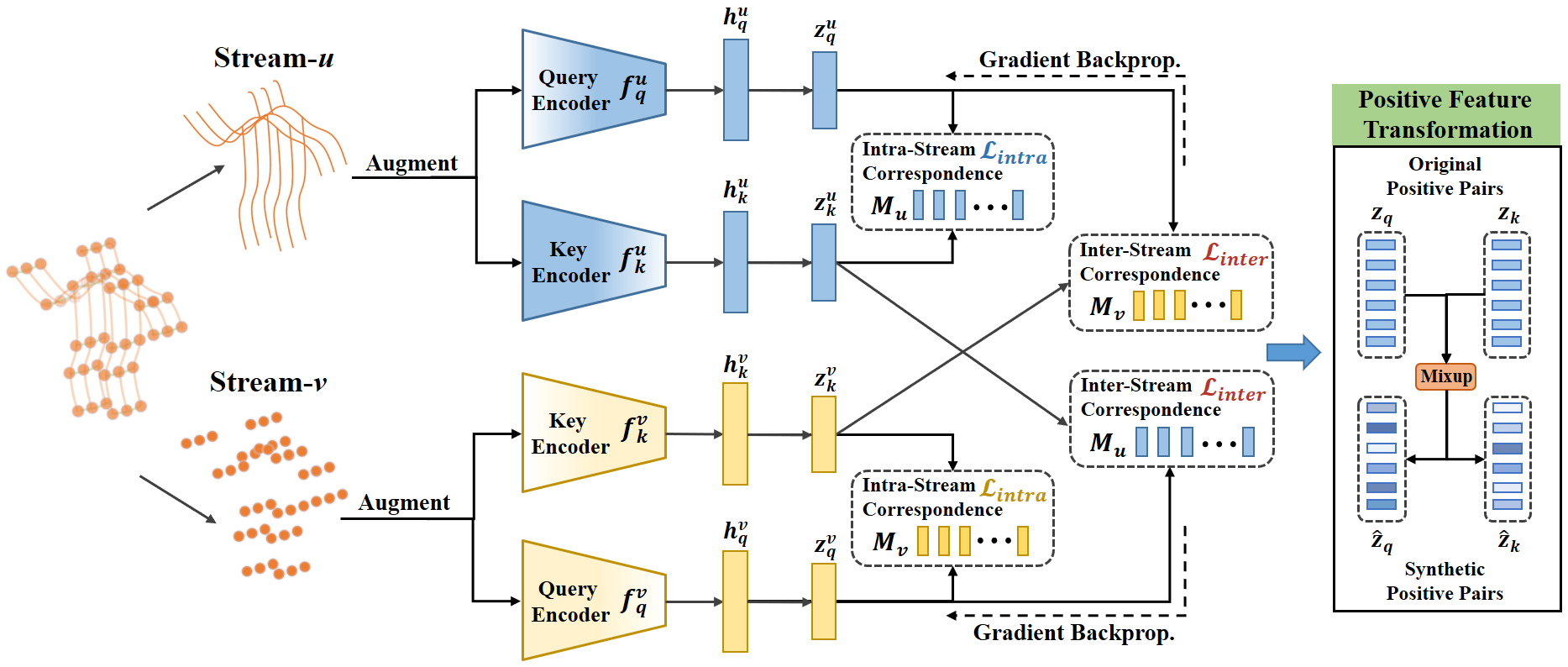}
	\caption{The Framework of our Cross-stream Contrastive Learning for Skeleton-based Action Recognition. The model takes skeleton data with different data streams as input, then extracts the query and key features. Next, intra-stream contrastive learning and inter-stream contrastive learning are adopted for training. Finally, the positive set is adjusted with feature transformation. Synthetic positive pairs are generated and features are updated to be more dissimilar (different color shades), which enable the model to learn more robust features for downstream tasks.}
	\label{fig2}
\end{figure*}

In this section, we first introduce preliminaries in Section \ref{sec-3-1}. Next, the pipeline of Intra-Stream Contrastive Learning for Skeleton Representation (IntraCLR) and Inter-Stream Contrastive Learning for Skeleton Representation (InterCLR) are described in Section \ref{sec-3-2}. Then, we demonstrate the Positive Feature Transformation (PFT) in Section \ref{sec-3-3}. Finally, more details of CSCLR are introduced in Section \ref{sec-3-4}. Briefly, we take a two-stream case (stream-$u$ and stream-$v$) as an example and illustrate the overview of our method in Figure \ref{fig2}.

\subsection{Preliminaries}\label{sec-3-1}
Initially, we are supposed to be given a skeleton sequence $x \in \mathop \mathbb{R}\nolimits^{T \times C \times V} $ as input, where $T$ is the temporal length, $C$ is the number of channels, and $V$ represents the number of keypoints of human body. Based on the raw skeleton data $x$, sample $x^u$ and sample $x^v$ in stream $u$ and $v$ are generated by method in Section \ref{sec-3-4}. Our aim is to train a skeleton encoder $f$ in self-supervised manner to be effectively transferable to down-stream tasks, $e.g.$ skeleton-based action recognition. Inspired by the great success of contrastive learning, we develop the self-supervised skeleton representation learning framework based on the recent advanced practice MoCo-v2.

The contrastive learning framework takes skeleton samples as input, and consists of three components: \emph{Data Augmentation, Feature Extraction,} and \emph{Model Training}. We will introduce the first two components in this subsection, and the model training of IntraCLR and InterCLR will be described in the next subsection. In the common contrastive learning pipeline, a skeleton sample is initially transformed into different augments, and these augments can be seen as positive sample pairs. Besides, other augmented samples transformed by different skeletons are regarded as negative samples. After extracting features of these samples, an InfoNCE \cite{oord2018representation} loss for instance discrimination is introduced for model training. Below, we take the skeleton data of stream-$u$ $x^u$ as example for elaboration.

\textbf{Data Augmentations.} We utilize a data augmentation module to transform the raw skeleton sequence $x^u$ into different augments $x_q^u$ and $x_k^u$, where $x_q^u$ represents the query sample, and $x_k^u$ represents the key sample. Due to the same source of these two augments, $x_q^u$ and $x_k^u$ are considered to be positive pairs for training. Both spatial and temporal augmentations are utilized for randomly transforming the input skeleton data, and the augmentation details are introduced in Section \ref{sec-4-2}.

\textbf{Encoder.} Two GCN-based skeleton encoders $f_q^u$ and $f_k^u$ are constructed for extracting deep features of $x_q^u$ and $x_k^u$ respectively, depicted as $\mathop h\nolimits_q^u  = \mathop f\nolimits_q^u \left( {\mathop x\nolimits_q^u; \theta_q^u } \right)$, $\mathop h\nolimits_k^u  = \mathop f\nolimits_k^u \left( {\mathop x\nolimits_k^u; \theta_k^u } \right)$. We use ST-GCN \cite{yan2018spatial} as the backbone network in practice. In the training process, momentum update is adopted when optimizing key encoder $\mathop f\nolimits_k^u$, and only parameters in query encoder $\mathop f\nolimits_q^u$ is updated with gradient backpropagation. Denoting the parameters of $\mathop f\nolimits_q^u$ as $\theta_q^u$ and those of $\mathop f\nolimits_k^u$ as $\theta_k^u$, $\theta_k^u$ is updated by:
\begin{flalign}
    \begin{split}
	   \theta_k^u \leftarrow m\theta_k^u + \left(1-m\right)\theta_q^u,
    \end{split}&
\end{flalign}
where $m\in\left[ {\left. {0,1} \right)} \right.$ is a momentum coefficient. Due to the momentum update, encoder $f_k^u$ is able to avoid the rapid change and maintain the consistency of key representation. In the following of skeleton encoder, a simple MLP layer $g$ with ReLU is utilized to project the hidden vector $h$ into a low-dimension feature space. For the query and key sample, the corresponding deep skeleton feature are computed as $z_q^u=\mathop g\nolimits_q^u \left( {\mathop h\nolimits_q^u} \right)$, $z_k^u=\mathop g\nolimits_k^u \left( {\mathop h\nolimits_k^u} \right)$. 

\subsection{Intra-Stream and Inter-Stream Contrastive Learning for Skeleton Representation}\label{sec-3-2}

In IntraCLR, only a single data stream is taken as input, the extracted feature of the two augments $z_q^u$ and $z_k^u$ are defined as positive pairs. As for negative pairs, similar to MoCo \cite{he2020momentum}, a dynamic memory bank $M_u = \mathop {\left\{ {\mathop m\nolimits_i^u } \right\}}\nolimits_{i = 1}^M $ is introduced to store negative samples. To enlarge the amount of negative samples, the memory bank consider a dictionary as a queue, where samples $x_k^u$ enqueue and dequeue in each iteration. To facilitate model training, contrastive loss is used for instance discrimination. If the query data $x_q^u$ is similar to its positive key $x_k^u$, the contrastive loss would be low. Otherwise, if the query data $x_q^u$ is similar to its negative key in the memory, this loss would be high. Thus, InfoNCE loss is used to pull the positive pairs close in the feature space, while push the negative pairs away. The loss of IntraCLR can be formulated as follows:
\begin{flalign}
    	&\mathcal{L}_{intra}^u  =  - \log \frac{{\exp \left( {\mathop z\nolimits_q^u  \cdot \mathop z\nolimits_k^u /\origtau } \right)}}{{\exp \left( {\mathop z\nolimits_q^u  \cdot \mathop z\nolimits_k^u /\origtau } \right) + \sum_{i=1}^M \exp \left( {\mathop z\nolimits_q^u  \cdot \mathop m\nolimits_i^u /\origtau } \right)} },&
\end{flalign}
where $\mathop m\nolimits_i^u \in M_u$, $\origtau$ is temperature coefficient, and we use dot product as our similarity function. 

In addition to IntraCLR, we demonstrate another contrastive objective InterCLR based on the way of inter-stream learning. 
To elaborate InterCLR, we first introduce another stream of skeleton data (stream-$v$) and extract GCN feature as IntraCLR, denoted as $x^v$ and $z^v$ respectively. Similarly, both query feature $z_q^v$ and key feature $z_k^v$ are generated for further training. When multiple streams of skeleton input are involved, the aim of contrastive learning would be stream-invariant. Specifically, the similarity of positive pairs from different streams ($e.g. $ stream-$u$ and stream-$v$) should be maximized, since they both correspond to the same raw skeleton data from the source. Compared with the paired sample within a single stream ($z_q^u$ and $z_k^u$), the inter-stream alignment brings in harder positive samples, and the increasing variance between inter-stream positive pairs ($z_q^u$ and $z_k^v$) will lead to more implicit information for contrastive learning. With sample pairs from different streams, the query encoder is able to learn more robust skeleton feature and capture spatial-temporal information of input skeleton sequence more effectively, which benefits the knowledge transfer to downstream tasks. After the feature extraction in both stream-$u$ and stream-$v$, the inter-stream loss can be formulated as:
{\setlength\belowdisplayskip{0.2pt}
\begin{flalign}
    \begin{split}
    	\mathcal{L}_{u\rightarrow v}  =  - \log \frac{{\exp \left( {\mathop z\nolimits_q^u  \cdot \mathop z\nolimits_k^v /\origtau } \right)}}{{\exp \left( {\mathop z\nolimits_q^u  \cdot \mathop z\nolimits_k^v /\origtau } \right) + \sum\nolimits_{i = 1}^M {\exp \left( {\mathop z\nolimits_q^u  \cdot \mathop m\nolimits_i^v /\origtau } \right)} }},
     \end{split}&
\end{flalign}}
\begin{flalign}
    	&\mathcal{L}_{v\rightarrow u}  =  - \log \frac{{\exp \left( {\mathop z\nolimits_q^v  \cdot \mathop z\nolimits_k^u /\origtau } \right)}}{{\exp \left( {\mathop z\nolimits_q^v  \cdot \mathop z\nolimits_k^u /\origtau } \right) + \sum\nolimits_{i = 1}^M {\exp \left( {\mathop z\nolimits_q^v  \cdot \mathop m\nolimits_i^u /\origtau } \right)} }}.&
\end{flalign}
As shown in equation 3 and equation 4, the bi-directional inter-stream loss considers features from different streams as input. Due to the discrepancy of inherent information in different streams, positive pairs with lower similarity are constructed in this way, which facilitates more robust representation learning. 

\textbf{Multi-stream Scenario.} When a set of streams $\mathcal{S}=\mathop {\left\{ {\mathop S\nolimits_i } \right\}}\nolimits_{i = 1}^{N_s}$ are taken as input, where $N_s$ represents the number of data streams, we need to combine both IntraCLR and InterCLR losses from all used streams. For IntraCLR, losses are combined as:
\begin{flalign}
    \begin{split}
    	\mathcal{L}_{intra}  = \sum\nolimits_{u \in \mathcal{S}} {\mathcal{L}_{intra}^u }.
    \end{split}&
\end{flalign}
Also, the objective of InterCLR can be computed as:
\begin{flalign}
    \begin{split}
    	\mathcal{L}_{inter} = \sum\nolimits_{u \in \mathcal{S}} {\sum\nolimits_{v \in \mathcal{S}} {\mathcal{L}_{u\rightarrow v} } },
    \end{split}&
\end{flalign}
where $u \ne v$.

Finally, we obtain the overall loss function as the combination of $\mathcal{L}_{intra}$ and $\mathcal{L}_{inter}$, where the former imposes invariance to data augmentations within a single stream, and the latter injects the correspondence between different streams. 
\begin{flalign}
    \begin{split}
    	\mathcal{L}=\mathcal{L}_{intra}+\mathcal{L}_{inter}.
    \end{split}&
\end{flalign}

\subsection{Positive Feature Transformation}\label{sec-3-3}
Except for generating positives with data augmentation pipeline, we propose to adjust the positive set by Positive Feature Transformation (PFT). PFT is designed to generate hard positives through feature extrapolations, which increases the variance of the positive set. In the guarantee of stable and smooth score distribution and gradient, the harder positives can be beneficial the transfer performance of downstream tasks \cite{guo2022contrastive,zhu2021improving}.

\begin{figure}[!t]
	\centering
	\includegraphics[width=8.6cm,height=3.45cm]{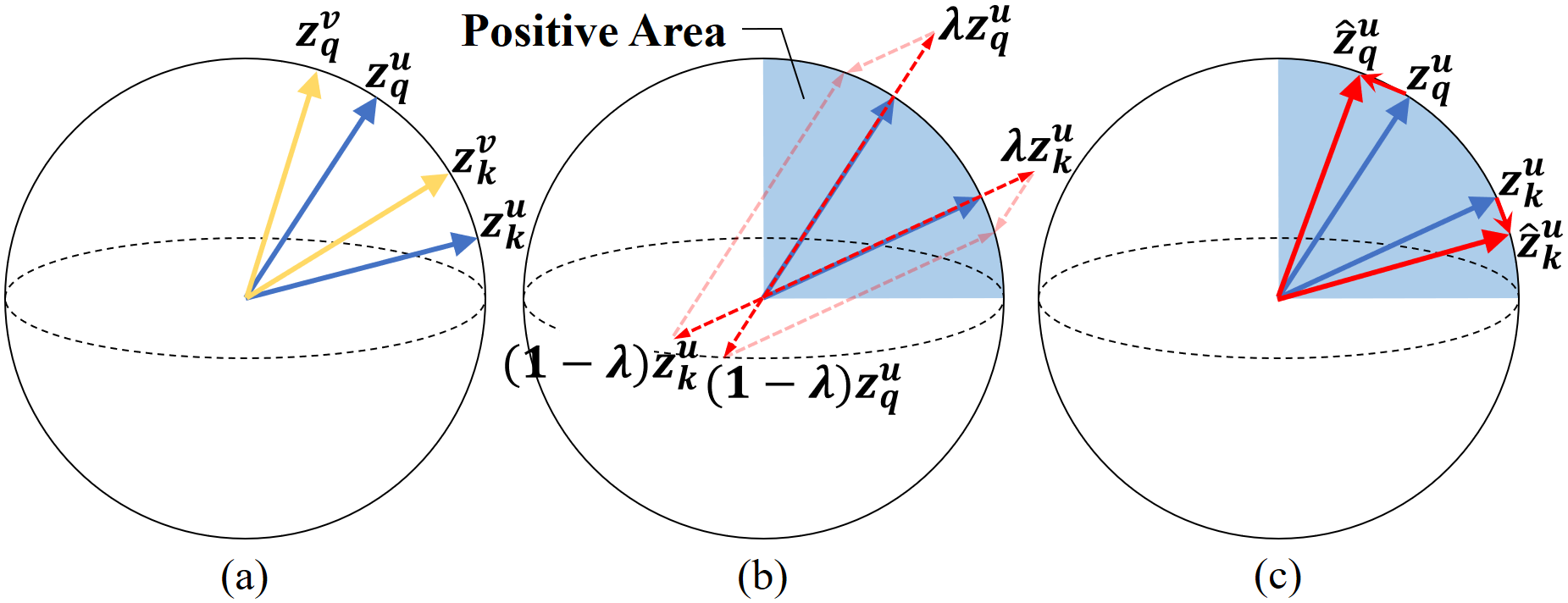}
	\caption{An illustration of our proposed PFT. After PFT, the original positive pairs are repelled, resulting in increasing view variance in positive set and more robust self-supervised representation learning.}
	\label{fig3}
\end{figure}

 Figure \ref{fig3} (a) shows two positive pairs in feature space. Similarly, we take the input of stream-$u$ as example. In order to encourage the model to generate more robust features, we manipulate the positive pair $\mathop z\nolimits_q^u$ and $\mathop z\nolimits_k^u$ to moderately increase the view variance between them. In other words, we add reasonable perturbations to the features and generate synthetic harder positives for contrastive learning. 

Inspired by the design of Mixup \cite{zhang2017mixup} and Manifold Mixup \cite{verma2018manifold}, we utilize weighted sum to integrate the positive pairs. As shown in Figure \ref{fig3} (b), the newly generated synthetic feature $\hat{z}^u_q$ and $\hat{z}^u_k$ are computed as:
\begin{flalign}
	\begin{split}
		\hat{z}^u_q = \lambda z^u_q + \left(1 - \lambda\right) z^u_k,  \\
		\hat{z}^u_k = \lambda z^u_k + \left(1 - \lambda\right) z^u_q.
	\end{split}& 
\end{flalign}
To address our aim of generating harder positives, we should guarantee that the similarity of $\hat{z}^u_q$ and $\hat{z}^u_k$ will be lower than that of $z^u_q$ and $z^u_k$. Formally, this means $\hat{z}^u_q \hat{z}^u_k \leq z^u_q z^u_k$ holds, when taking dot product as similarity metric. Based on Equation (8), we can obtain:
\begin{flalign}
    \begin{split}
    	\hat{z}^u_q \hat{z}^u_k 
    	& = \left[\lambda z^u_q + \left(1 - \lambda\right) z^u_k\right] \left[\lambda z^u_k + \left(1 - \lambda\right) z^u_q\right] \\
    	& = 2\lambda\left(1-\lambda\right)\left(1-z^u_q z^u_k\right) + z^u_q z^u_k,
    \end{split}&
\end{flalign}
where $0 \leq z^u_q z^u_k < 1$. From the Equation (9), we can observe that if $\lambda \geq 1$, then $2\lambda\left(1-\lambda\right) \leq 0$ and $\hat{z}^u_q \hat{z}^u_k \leq z^u_q z^u_k$ holds. Thus, we formulate a beta-like distribution $F\left(\alpha, \mu\right)$ as follows and sample the value of $\lambda$ from it, $\lambda \sim F\left(\alpha, \mu\right)$.
\begin{flalign}
    \begin{split}
    	F\left(\alpha, \mu\right) = Beta\left(\alpha, \alpha\right)\times \mu + 1,
    \end{split}&
\end{flalign}
where $\alpha$ is the parameter of $Beta$ distribution and $\mu$ is a parameter to limits the value of $\lambda$. Consequently, this makes $\lambda \in [1, 1 + \mu)$. As shown in Figure \ref{fig3} (c), a subtle direction shift is applied to the original positive pairs $z^u_q$ and $z^u_k$. After that, the angle between transformed feature $\hat{z}^u_q$ and $\hat{z}^u_k$ are larger than the angle between original feature $z^u_q$ and $z^u_k$, indicating the larger variance in the positive set.

\begin{figure}[t]
	\centering
	\includegraphics[width=7.0cm,height=5.2cm]{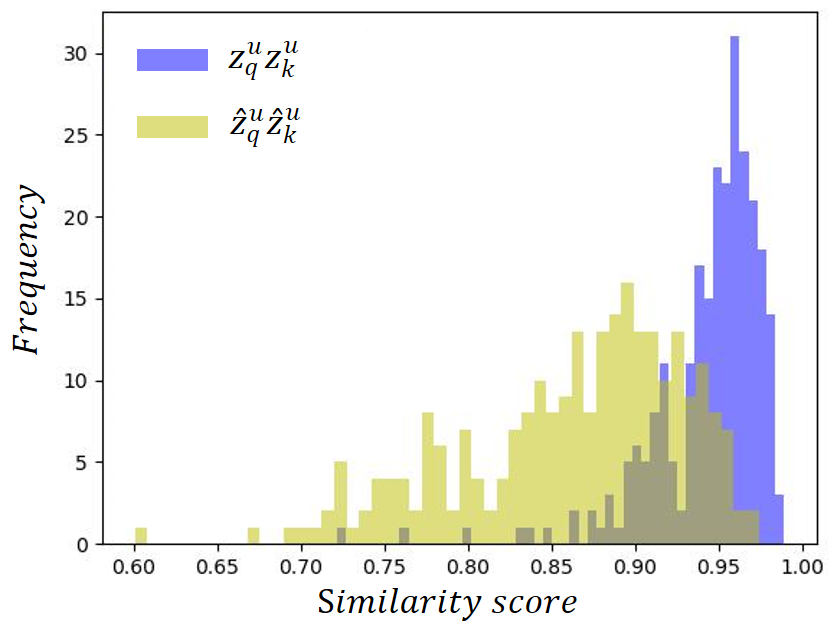}
	\caption{The comparison of similarity score distribution $z^u_q z^u_k$ and $\hat{z}^u_q \hat{z}^u_k$. After PFT, the hard positives are generated, and the variation of similarity in the positive set is evidently increased.}
	\label{fig4}
\end{figure}

In contrastive learning, the similarity scores of positive pairs are defined to be non-negative, while that of negative pairs are non-positive. Intuitively, the angle between transformed positive feature should be $90^\circ$ at most, which corresponds to the positive area in Figure \ref{fig3}. In case that the similarity score of a few positive pairs are excessively decreased to negative values after the feature transformation, we do not adopt PFT on those pairs. Figure \ref{fig4} shows the similarity score distribution of $z^u_q z^u_k$ and $\hat{z}^u_q \hat{z}^u_k$ in a randomly-selected mini-batch. First, the similarity scores between positive pairs become lower, and the variances in the positive set get higher after PFT. Second, there is no negative value of $\hat{z}^u_q \hat{z}^u_k$ after adopting the strategy above, which averts the contradiction with the definition of positive sample pairs.


%

\subsection{Model Details}\label{sec-3-4}

\textbf{Different Data Streams of 3D Skeleton.} Except for the 3D joint coordinates $x \in \mathop \mathbb{R}\nolimits^{T \times C \times V} $, multiple data streams of skeleton are generated as \cite{shi2019skeleton, shi2019two}, $e.g.$ bone, motion. Bone is represented as a vector pointing to its target joint from its source joint ($x_{:,:,v_2}-x_{:,:,v_1}$), which contains not only the length information, but also the direction information. And motion is obtained by computing the displacement between adjacent frames ($x_{t_2,:,:}-x_{t_1,:,:}$), which brings in the temporal shift information. In this paper, we use three streams: joint, bone and motion in experiments.

\textbf{Nearest Neighbors Mining (NNM).} Similar to AimCLR \cite{guo2022contrastive}, we also apply NNM and take the mined samples in memory queue as positives for training. Although these selected neighbors are stored in the memory queue, they are quite similar with query $z^u_q$. With these mined positives, the intra-stream objective of contrastive learning could be formulated as follows:
\begin{flalign}
    \begin{split}
    	\mathcal{L}_{N-intra}^u  =  - \log \frac{{\exp \left( {\mathop z\nolimits_q^u  \cdot \mathop z\nolimits_k^u /\origtau } \right) + \sum\nolimits_{i \in N_+^u} {\exp \left( {\mathop z\nolimits_q^u  \cdot \mathop m\nolimits_i^u /\origtau } \right)}}}{{\exp \left( {\mathop z\nolimits_q^u  \cdot \mathop z\nolimits_k^u /\origtau } \right) + \sum\nolimits_{i = 1}^M {\exp \left( {\mathop z\nolimits_q^u  \cdot \mathop m\nolimits_i^u /\origtau } \right)} }},
    \end{split}
\end{flalign}
where $N^u_+$ is the index set of selected neighbors of $z^u_q$ in $M_u$.

\textbf{Multi-stage Training Strategy.} In the earlier epochs, the unstable model is not able to provide effective features for NNM and PFT, so we perform multi-stage training for CSCLR. The whole training process is divided into three stages: Basic training, Basic training + NNM, Basic training + NNM + PFT. In the first stage, we train our model with the combination of $\mathcal{L}_{intra}$ and $\mathcal{L}_{inter}$. Next, we adopt NNM to increase the number of positive pairs with reliable neighbors. Finally, we adopt PFT to generate hard positives and improve the robustness of representation learning.

\section{Experiments and results}\label{sec-4}
In this section, we first introduce datasets in Section \ref{sec-4-1}. Next, the experimental settings for a fair comparison are described in Section \ref{sec-4-2}. Then, the comparison with the state-of-the-art methods are shown in Section \ref{sec-4-3}. Finally, we demonstrate the results of ablation study in Section \ref{sec-4-4}.

\subsection{Dataset and Evaluation Metric} \label{sec-4-1}
\textbf{NTU-RGB+D 60.} The dataset consists of 56,578 skeleton sequences with 3D joint coordinate for skeleton-based action recognition, and these sequences are labeled with 60 action categories. In each skeleton graph, 25 joints throughout the body are set as nodes, and their 3D coordinates are obtained by Kinect V2 cameras. There are two suggested protocols. The first one is Cross-Subject (xsub), where skeleton sequences in training set and validation set are collected from different subjects. And the other one is Cross-View (xview), where skeleton sequences in training set and validation set are collected from different camera views.

\textbf{NTU-RGB+D 120.} The dataset \cite{liu2019ntu} is an extension of NTU-60, and this large-scale benchmark dataset contains 113,945 skeleton sequences in 120 action categories. There are two suggested protocols. The first one is Cross-Subject (xsub), where skeleton sequences in training set and validation set are collected from different subjects. And the other one is Cross-Setup (xsetup), where skeleton sequences in training set and validation set are collected from different setup IDs.

\textbf{PKU-MMD Dataset.} It contains almost 20,000 action sequences covering 51 action classes. It consists of two subsets. Part I is an easier version for action recognition, while part II is more challenging with more noise caused by view variation. We conduct experiments under the cross-subject protocol on the two subsets.

\subsection{Experiment Settings} \label{sec-4-2}
All the experiments are conducted on PyTorch \cite{paszke2017automatic}. For data pre-processing, we follow AimCLR \cite{guo2022contrastive}, SkeletonCLR \cite{li20213d} and CrosSCLR \cite{li20213d} for a fair comparison. 

\textbf{Data Augmentation.} We use the same data augmentations as AimCLR \cite{guo2022contrastive}, which includes \emph{Normal Augmentations} and \emph{Extreme Augmentations}. Here, we focus on the cross-stream contrastive learning, so we only introduce these augmentations briefly.

The Normal Augmentations consists of one spatial augmentation \emph{Shear} and one temporal augmentation \emph{Crop}. The Extreme Augmentations includes four spatial augmentations: \emph{Shear}, \emph{Spatial Flip}, \emph{Rotate}, \emph{Axis Mask} and two temporal augmentations: \emph{Crop}, \emph{Temporal Flip} and two spatio-temporal augmentations: \emph{Gaussian Noise}, \emph{Gaussian Blur}.

\textbf{Self-supervised Pre-training.} We use the same setting for contrastive learning as that in AimCLR, SkeletonCLR and CrosSCLR, the batch size is set to 128, and the size of memory queue is 32768. In the training process, we use SGD with momentum (0.9) and weight decay (0.0001). For the multi-stage training strategy mentioned in Section \ref{sec-3-4}, the model is trained for 150 epochs in the basic training stage, and then trained for 150 epochs with NNM involved. Finally, the model is trained for 200 epochs with PFT involved. The learning rate is set as 0.1, and decreases to 0.01 at epoch 250. In NNM, we only take the top-1 nearest neighbor in the memory queue as positive sample. For a fair comparison, we use the weights of [0:6; 0:6; 0:4] for fusing the three-stream predictions like other multi-stream GCN methods.

\textbf{Linear Evaluation Protocol.} To verify our model, we adopt linear evaluation for the action recognition task. To be specific, we train a linear classifier (a fully-connected layer followed by a softmax layer), and the encoder is frozen in optimization. The model is trained for 100 epochs with learning rate 3.0 (decrease to 0.3 at epoch 80).

\textbf{Finetuned Evaluation Protocol.} We append a linear classifier to the trained encoder, and the whole model (encoder and classifier) is optimized for the action recognition task, to compare it with fully supervised methods. The model is trained for 100 epochs with
learning rate 0.1 and weight decay 0.0001.

\textbf{Semi-supervised Evaluation Protocol.} The encoder is pre-trained with all data in self-supervised manner, and the whole model is finetuned with only 1\% or 10\% randomly selected labeled data.

\textbf{Competitors.} In addition to the self-comparison experiments, we compare the proposed CSCLR method with the state-of-the-art methods, including single-stream and multi-stream works. The single-stream setting includes LongT GAN \cite{zheng2018unsupervised}, MS$^2$L \cite{lin2020ms2l}, P\&C \cite{su2020predict}, SeBiReNet \cite{nie2020unsupervised}, AS-CAL \cite{rao2021augmented}, ST-CL \cite{gao2021efficient} and MG-AL \cite{9841515}, the performance of joint stream in SkeletonCLR \cite{li20213d}, CrosSCLR \cite{li20213d} and AimCLR \cite{guo2022contrastive} are also demonstrated for single-stream comparison. For the multi-stream setting, we mainly include 3s-SkeletonCLR, 3s-CrosSCLR and 3s-AimCLR, where predictions of three streams (joint, bone and motion) are fused.

\begin{table}[h]
    
    \caption{\textrm{Linear evaluation accuracy comparisons with the state-of-the-art methods on NTU-60 dataset. “3s” represents three stream fusion.``$\dagger$'' means using cross-stream knowledge mining strategy proposed in 3s-CrosSCLR. The best results is in bold face.}}
    \label{tab-3}
    \renewcommand{\arraystretch}{1.2}
    \scalebox{0.9}{
    \begin{tabular}{ccccccc}
    \hline
    \multirow{2}{*}{Category}              & \multirow{2}{*}{Method} & \multirow{2}{*}{Year} & \multicolumn{2}{c}{NTU-60 (\%)}              \\
    \cline{4-5} 
    &                         &                       & xsub                 & xview                \\ \hline
    \specialrule{0em}{1.2pt}{1.2pt}
    \multirow{10}{*}{\begin{tabular}[c]{@{}c@{}}Single\\Stream\end{tabular}} & LongT GAN              &  AAAI'18                     &  39.1                    &                      48.1\\
    & MS$^2$L                    & MM'20                      & 52.6                     &                      -\\
    & P\&C                    & CVPR'20                      & 50.7                      &                      76.3\\
    & SeBiReNet               & ECCV'20                      & -                     &                      79.7\\
    & SkeletonCLR             & CVPR'21                      & 68.3                     &                      76.4\\
    & AS-CAL                  & Info. Sci.'21                      & 58.5                    &                      64.8\\
    & ST-CL                  & TMM'21                      & 68.1                    &                      69.4\\
    & MG-AL				   & TCSVT'22            & 64.7     &       68.0\\
    & AimCLR               & AAAI'22             & 74.3     &       79.7\\
    & \bfseries CSCLR (Ours)        & - &  \bfseries75.7& \bfseries81.3 \\ \cline{1-5} 
    \specialrule{0em}{1.2pt}{1.2pt}
    \multirow{5}{*}{\begin{tabular}[c]{@{}c@{}}Multiple\\Stream\end{tabular}}  & 3s-SkeletonCLR          & CVPR'21                      & 75.0                    &                      79.8\\
    & 3s-CrosSCLR          & CVPR'21             & 77.8     &       83.4\\
    & 3s-AimCLR$^\dagger$               & AAAI'22             & 78.6     &       82.6\\
    & 3s-AimCLR               & AAAI'22             & 78.9     &       83.8\\
    & \bfseries 3s-CSCLR (Ours)              & -       &             \bfseries80.1&   \bfseries85.2   \\
    \cline{1-5} 
    \end{tabular}
}
\end{table}

\begin{table}[]
	\caption{\textrm{Linear evaluation accuracy comparisons with the state-of-the-art methods on NTU-120 dataset. The best results is in bold face.}}
	\label{tab-4}
	\renewcommand{\arraystretch}{1.2}
        \scalebox{0.9}{
	\begin{tabular}{ccccccc}
		
		\hline
		\multirow{2}{*}{\textrm{Category}}              & \multirow{2}{*}{\textrm{Method}} & \multirow{2}{*}{\textrm{Year}} & \multicolumn{2}{c}{\textrm{NTU-120 (\%)}}              \\
        \cline{4-5}
		&                         &                       & \textrm{xsub}                 & \textrm{xsetup }               \\ \hline
		\specialrule{0em}{1.2pt}{1.2pt}
		\multirow{7}{*}{\begin{tabular}[c]{@{}c@{}}\textrm{Single}\\\textrm{Stream}\end{tabular}}
		& \textrm{P\&C}                    & \textrm{CVPR'20}                      & \textrm{42.7}                      &                      \textrm{41.7}\\ 
		& \textrm{AS-CAL}                  & \textrm{Info. Sci.'21}                      & \textrm{48.6}                    &                      \textrm{49.2}\\
		& \textrm{ST-CL}                  & \textrm{TMM'21}                      & \textrm{54.2}                    &                      \textrm{55.6}\\
        & \textrm{SkeletonCLR}             & \textrm{CVPR'21}                      & \textrm{56.8}                     &                      \textrm{55.9}\\
		& \textrm{MG-AL}                  & \textrm{TCSVT'22}                      & \textrm{46.2}                    &                      \textrm{49.5}\\
        & \textrm{AimCLR}                 & \textrm{AAAI'22}             & \textrm{63.4}     &       \textrm{63.4}\\
		& \bfseries \textrm{CSCLR (Ours)}        & \textrm{-} &\bfseries\textrm{64.5}  & \bfseries\textrm{64.3} \\ \cline{1-5} 
		\specialrule{0em}{1.2pt}{1.2pt}
		\multirow{5}{*}{\begin{tabular}[c]{@{}c@{}}\textrm{Multiple}\\\textrm{Stream}\end{tabular}}  & \textrm{3s-SkeletonCLR}          & \textrm{CVPR'21}                      & \textrm{60.7}                    &                      \textrm{62.6}\\
		& \textrm{3s-CrosSCLR}          & \textrm{CVPR'21}                      & \textrm{67.9}                    &                      \textrm{66.7}\\
		& \textrm{3s-AimCLR}$^\dagger$               & \textrm{AAAI'22}             & \textrm{68.0}     &       \textrm{68.7}\\
		& \textrm{3s-AimCLR}               & \textrm{AAAI'22}             & \textrm{68.2}     &       \textrm{68.8}\\
		& \bfseries \textrm{3s-CSCLR (Ours)}              & \textrm{-}       &  \bfseries\textrm{69.2}           & \bfseries\textrm{70.2}    \\ \cline{1-5}
	\end{tabular}
    }
\end{table}

\begin{table}[]
	\caption{\textrm{Linear evaluation accuracy comparisons with the state-of-the-art methods on PKU-MMD dataset. The best results is in bold face.}}
	\label{tab-5}
	\renewcommand{\arraystretch}{1.2}
    \scalebox{0.9}{
	\begin{tabular}{ccccccc}
		\hline
		\specialrule{0em}{1.2pt}{1.2pt}
		\multirow{2}{*}{Category}              & \multirow{2}{*}{Method} & \multirow{2}{*}{Year} & \multicolumn{2}{c}{PKU-MMD (\%)}              \\
        \cline{4-5}
		& & & part \uppercase\expandafter{\romannumeral1}  & part \uppercase\expandafter{\romannumeral2}               \\ \hline
		\multirow{5}{*}{\begin{tabular}[c]{@{}c@{}}Single\\Stream\end{tabular}} 
		& LongT GAN               &  AAAI'18                     &  67.7                    &                      26.0\\
		& MS$^2$L                    & MM'20                      & 64.9                     &                      27.6\\
        & SkeletonCLR             & CVPR'21                      & 80.9                     &                      -\\
        & AimCLR               & AAAI'22             & 83.4     &       -\\
		& \bfseries CSCLR (Ours)        & - & \bfseries85.3 & \bfseries31.8 \\ \cline{1-5} 
		\specialrule{0em}{1.2pt}{1.2pt}
		\multirow{4}{*}{\begin{tabular}[c]{@{}c@{}}Multiple\\Stream\end{tabular}}  & 3s-SkeletonCLR          & CVPR'21                      & 84.9                    &                      21.2\\
		&3s-AimCLR$^\dagger$                    & AAAI'22                      & 87.4                     &                      39.5\\
		&3s-AimCLR                    & AAAI'22                      & 87.8                     &                      38.5\\
		&\bfseries 3s-CSCLR (Ours)              & -       &\bfseries89.3             &\bfseries45.1      \\ \cline{1-5} 
	\end{tabular}
    }
\end{table}

\subsection{Comparison with the State-of-the-art} \label{sec-4-3}

We compare the proposed method with state-of-the-art methods on NTU-60, NTU-120 and PKU-MMD datasets with the corresponding protocol.

\textbf{Linear Evaluation Protocol Results.}
Results on NTU-60, NTU-120 and PKU-MMD are shown in Table \ref{tab-3}, Table \ref{tab-4} and Table \ref{tab-5} respectively. For a fair comparison, we report both single-stream (joint) and multi-stream fusion results. 

As can be seen from Table \ref{tab-3}, the result of CSCLR single stream significantly outperforms the previous methods with joint-stream data on NTU-60. For the fusion results, our CSCLR still maintains the advantage over the previous methods under both x-sub and x-view protocols. Compared with 3s-CrosSCLR, the CSCLR gains 2.3\% and 1.8\% accuracy improvements under x-sub and x-view respectively. When comparing with 3s-AimCLR, CSCLR outperforms it by 1.2\% and 1.4\% respectively. It is worth mentioning that the performance of 3s-AimCLR are decreased when using cross-stream knowledge mining strategy in 3s-CrosSCLR, while better performance are boosted when adopting the proposed CSCLR. The improved performance proves that our method is able to take the advantage of correlating information between different data streams and inherit knowledge in positive set.

In Table \ref{tab-4}, our CSCLR also enjoys evident accuracy improvement over existing self-supervised methods on NTU-120. Under single-stream setting, CSCLR-joint significantly surpasses the advanced ST-CL (64.5\% vs 54.2\% on xsub and 64.3\% vs 55.6\% on xsetup). For fusion results, the accuracy of CSCLR is also higher than that of 3s-AimCLR (69.2\% vs 68.2\% on xsub and 70.2\% vs 68.8\% on xsetup). The advantages in Table \ref{tab-4} indicates the competitiveness of CSCLR when conducting experiments on large-scale datasets.

As demonstrated in Table \ref{tab-5}, the proposed CSCLR outperforms all compared state-of-the-art self-supervised methods on PKU-MMD. Specifically in part I and part II subsets, our method achieves accuracy of 85.3\% and 31.8\% with single data stream, and gains accuracy of 89.3\% and 45.1\% respectively with multi-stream fusion. Notably, CSCLR leads 3s-AimCLR 6.6\% under the part II subset, which indicates that the trained CSCLR model significantly benefits the downstream task on more challenging and noisy dataset.

\begin{table}[h]
	\caption{\textrm{Finetuned evaluation accuracy comparisons with the state-of-the-art methods on NTU-60 and NTU-120 dataset. The best results is in bold face.}}
	\label{tab-7}
	\renewcommand{\arraystretch}{1.2}
	\centering
        \scalebox{0.9}{
	\begin{tabular}{cccccc}
		
		\hline
		\specialrule{0em}{1.2pt}{1.2pt}
		\multirow{2}{*}{\textrm{Method}} & \multirow{2}{*}\textrm{{Year}} & \multicolumn{2}{c}{\textrm{NTU-60 (\%)}}  & \multicolumn{2}{c}{\textrm{NTU-120 (\%)}}            \\
        \cmidrule(r){3-4}
        \cmidrule(r){5-6}
		&  & \textrm{xsub}  & \textrm{xview} & \textrm{xsub} & \textrm{xsetup}               \\ \hline
		\textrm{SkeletonCLR}               &  \textrm{CVPR'21}                     &  \textrm{82.2}                    &                      \textrm{88.9}  & \textrm{73.6} & \textrm{75.3}\\
		\textrm{AimCLR}                    & \textrm{AAAI'22}                      & \textrm{83.0}                    &                      \textrm{89.2} & \textrm{76.4} & \textrm{76.7}\\
		\bfseries \textrm{CSCLR (Ours)}			        & - &\bfseries \textrm{84.7}  &\bfseries\textrm{90.4} &\bfseries \textrm{76.9}  & \bfseries \textrm{77.6} \\ \hline
		\specialrule{0em}{1.2pt}{1.2pt}
		\textrm{3s-ST-GCN}                    & \textrm{AAAI'18}                      & \textrm{85.2}                     &                      \textrm{91.4} & \textrm{77.2} & \textrm{77.1}\\
		\textrm{3s-CrosSCLR}          & \textrm{CVPR'21}                      & \textrm{86.2}                    &                      \textrm{92.5} & \textrm{80.5} & \textrm{80.4}\\
		\textrm{3s-AimCLR}                    & \textrm{AAAI'22}                      & \textrm{86.9}                     &                      \textrm{92.8} & \textrm{80.1} & \textrm{80.9}\\
		\bfseries \textrm{3s-CSCLR (Ours)}              & \textrm{-}       & \bfseries\textrm{87.0}            & \bfseries\textrm{93.6} & \bfseries\textrm{80.2} & \bfseries\textrm{81.6}       \\ \hline 
	\end{tabular}
    }
\end{table}

\textbf{Finetuned Evaluation Protocol Results.} We compare the CSCLR with other methods under finetuned evaluation protocol, and both single-stream and multi-stream fusion results are represented in Table \ref{tab-7}. All the competitors are developed based on the same encoder, ST-GCN. Similar to results in linear evaluation protocol, our CSCLR outperform ST-GCN and other self-supervised methods on both NTU-60 and NTU-120 dataset, indicating the efficacy of our proposed method.

\begin{figure*}[]
	\centering
    \begin{subfigure}{0.325\linewidth}
		\includegraphics[width=1\textwidth]{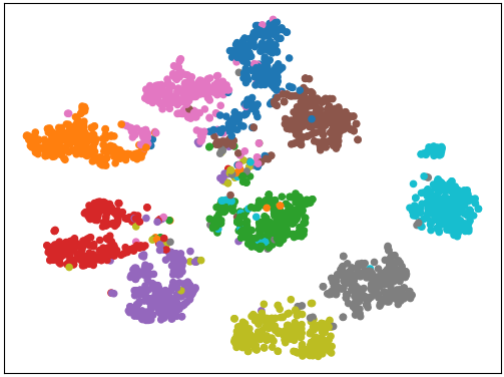}
		\caption{AimCLR-joint}
		\label{}
	\end{subfigure}\hfill
	\begin{subfigure}{0.325\linewidth}
		\includegraphics[width=1\textwidth]{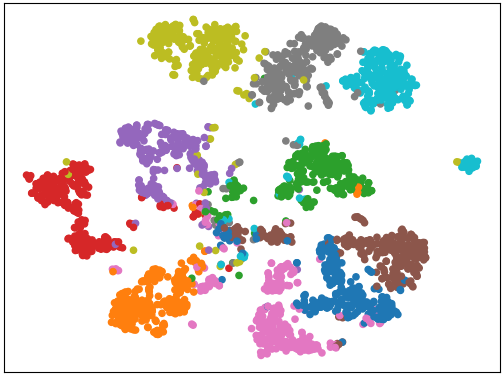}
		\caption{AimCLR-bone}
		\label{}
	\end{subfigure}\hfill
	\begin{subfigure}{0.325\linewidth}
		\includegraphics[width=1\textwidth]{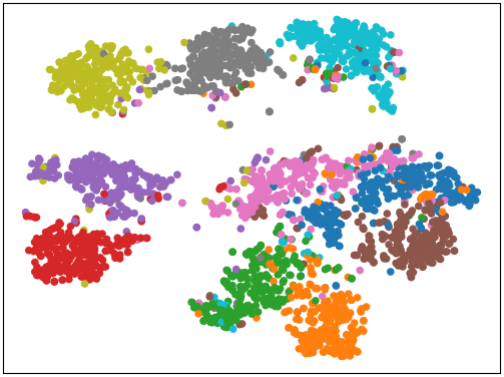}
		\caption{AimCLR-motion}
		\label{}
	\end{subfigure}\\
	
	\begin{subfigure}{0.325\linewidth}
		\includegraphics[width=1\textwidth]{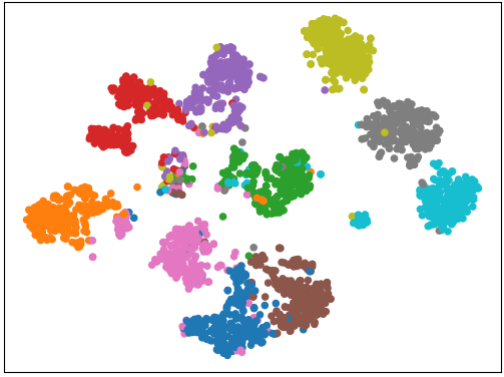}
		\caption{CSCLR-joint}
		\label{}
	\end{subfigure}\hfill
	\begin{subfigure}{0.325\linewidth}
		\includegraphics[width=1\textwidth]{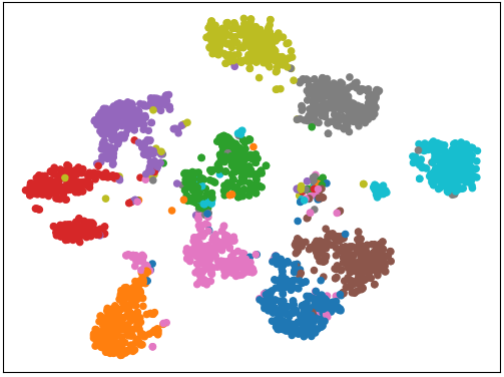}
		\caption{CSCLR-bone}
		\label{}
	\end{subfigure}\hfill
	\begin{subfigure}{0.325\linewidth}
		\includegraphics[width=1\textwidth]{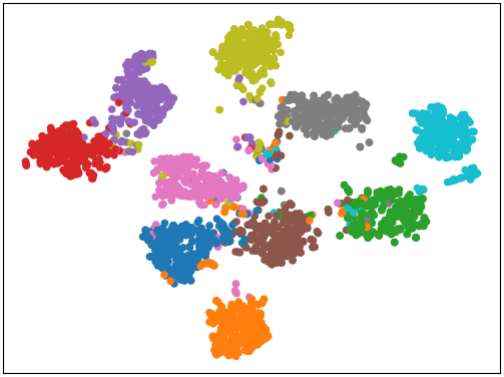}
		\caption{CSCLR-motion}
		\label{}
	\end{subfigure}
	\caption{The t-SNE visualization of embedding distributions on NTU-60-xview dataset. Embeddings from 10 categories are sampled and visualized with different colors. AimCLR and our CSCLR results in all used streams (joint, bone, motion) are showed from left to right respectively.}
	\label{fig6}
\end{figure*}

\textbf{Semi-supervised Evaluation Protocol Results.}
To further evaluate the self-supervised model, we demonstrate the semi-supervised evaluation results with 1\% and 10\% labeled data on NTU-60 and PKU-MMD. As shown in Table \ref{tab-6}, our CSCLR performs better than the previous competitors in most cases. The improvements on semi-supervised evaluation also indicate that CSCLR has positive effect of increasing the robustness of representation learning with extremely limited labeled data.

\textbf{Qualitative Results.} To evaluate our proposed method qualitatively, we apply t-SNE \cite{van2008visualizing} with fixed settings to show the embedding distribution of the advanced AimCLR and our CSCLR on NTU-60-xview dataset. Figure \ref{fig6} shows the comparison between results of CSCLR and AimCLR, it can be seen that the features from CSCLR performs better to enlarge the margin among the inter-class samples and reduce the intra-class distance at the same time, which demonstrates CSCLR is conducive to learn more discriminative features when transferring to downstream tasks.

\begin{table*}[h]
	\caption{\textrm{Semi-supervised evaluation results on NTU-60 and PKU-MMD dataset. 1\% and 10\% indicate that we use 1\% and 10\% labeled data for training the action recognition model. The best results is in bold face.}}
	\label{tab-6}
	\renewcommand{\arraystretch}{1.2}
	\centering
    \textrm{
	\begin{tabular}{lcccccccccccc}
		\hline
		\specialrule{0em}{1.2pt}{1.2pt}
		\multirow{2}{*}{Method} & \multirow{2}{*}{Year} & \multicolumn{2}{c}{NTU-60-xsub}  & \multicolumn{2}{c}{NTU-60-xview} & \multicolumn{2}{c}{PKU-MMD I} &\multicolumn{2}{c}{PKU-MMD II} \\ \cmidrule(r){3-4} \cmidrule(r){5-6} \cmidrule(r){7-8} \cmidrule(r){9-10}
		&                       & 1\%                    & 10\%   & 1\%                     & 10\%   & 1\%                      & 10\%  & 1\%                      & 10\%  \\ \hline
		\specialrule{0em}{1.2pt}{1.2pt}
		LongT GAN               &AAAI'18                       & 35.2                       & 62.0 &      -                        & -       & 35.8                         & 69.5   &12.4 & 25.7     \\
		MS$^2$L                  & MM'20                      & 33.1                       & 65.2       &  -          & -      & 36.4                         & 70.3    &13.0 & 26.1     \\
		ISC                  & MM'21                      & 35.7                        & 65.9  &      38.1                        & 72.5       & 37.7                         & 72.1   & - & -      \\
		3s-CrosSCLR                  & CVPR'21                      & 51.1                        & 74.4          & 50.0      & 77.8               & 49.7          & 82.9    & 10.2 & 28.6     \\
		3s-Colorization                  & ICCV'21                      & 48.3                        & 71.7          & 52.5      & 78.9               & -          & -     & - & -    \\
		3s-AimCLR                  & AAAI'22                      & 54.8                        & 78.2          & 54.3      & 81.6               & 57.5          & 86.1    & 15.1 & 33.4    \\
		\bfseries 3s-CSCLR (Ours)        & -  &\bfseries55.4 &\bfseries78.6          & \bfseries57.1    &\bfseries81.8        &\bfseries58.0      & {\bfseries86.1}    & \bfseries18.0   & \bfseries38.6    \\ \hline
	\end{tabular}
    }
\end{table*}

\subsection{Ablation Study} \label{sec-4-4}
To verify the effectiveness of the proposed method, we conduct ablation studies and provide the experimental results as follows. We follow the self-supervised pre-training, the linear evaluation and finetuned evaluation protocol.

\begin{table*}[t]
	\caption{\textrm{Linear evaluation results compared with AimCLR on NTU-60, PKU-MMD I and NTU-120 dataset. ``Repro.'' means the reproduced results. The best results is in bold face.}}
	\renewcommand{\arraystretch}{1.2}
	\centering
	\label{tab-2}
    \textrm{
	\begin{tabular}{lccccccccccc}
		\hline
		\specialrule{0em}{1.2pt}{1.2pt}
		\multirow{2}{*}{Method} & \multirow{2}{*}{Stream} & \multicolumn{2}{c}{NTU-60(\%)} &\multicolumn{2}{c}{PKU-MMD(\%)}&  \multicolumn{2}{c}{NTU-120(\%)} \\ \cmidrule(r){3-4} \cmidrule(r){5-6} \cmidrule(r){7-8}
		&                       & xsub                    & xview   & part I & part II                  & xsub                      & xsetup    \\ \hline
		\specialrule{0em}{1.2pt}{1.2pt}
		AimCLR                  & joint                   & 74.3         & 79.7         & 83.4    &-& 63.4         & 63.4          \\ 
		AimCLR (Repro.)                 & joint                   & 74.5         & 77.4         &83.1     & 30.2 & 63.3          & 63.9          \\ 
		\bfseries CSCLR (Ours)                    & joint                   & \bfseries75.7         & \bfseries81.3         & \bfseries85.3     & \bfseries31.8  & \bfseries64.5             & \bfseries64.3              \\ \hline
		\specialrule{0em}{1.2pt}{1.2pt}
		AimCLR                  & bone                    & 73.2         & 77.0         & 82.0    &-& 62.9         & 63.4          \\  
		AimCLR (Repro.)                 & bone                   & 72.3         & 76.8         & 81.9    & 28.2 & 61.2         & 64.7          \\ 
		\bfseries CSCLR (Ours)                    & bone                    & \bfseries76.3             & \bfseries80.1         & \bfseries83.3        & \bfseries 39.2 &  \bfseries64.8             & \bfseries65.1              \\ \hline
		\specialrule{0em}{1.2pt}{1.2pt}
		AimCLR                  & motion                  & 66.8         & 70.6         & 72.0    &-& 57.3         & 54.4          \\  
		AimCLR (Repro.)                 & motion                   & 64.5         & 71.2         &72.2     & 29.9  & 53.5         & 55.9          \\ 
		\bfseries CSCLR (Ours)                    & motion                  & \bfseries72.3             & \bfseries76.7        & \bfseries81.2        & \bfseries31.0  & \bfseries59.2             & \bfseries59.6              \\ \hline
		\specialrule{0em}{1.2pt}{1.2pt}
		3s-AimCLR                  & joint+bone+motion       & 78.9         & 83.8         & 87.8    & 38.5 & 68.2         & 68.8          \\ 
		3s-AimCLR (Repro.)                 & joint+bone+motion   & 78.6         & 83.2         & 86.9    & 35.0   &  67.9         & 69.5          \\ 
		\bfseries 3s-CSCLR (Ours)                    & joint+bone+motion       & \bfseries80.1             & \bfseries85.2         & \bfseries89.3    & \bfseries45.1   &  \bfseries69.2        &  \bfseries70.2             \\ \hline
	\end{tabular}
    }
\end{table*}

\textbf{The Effectiveness of CSCLR.} To verify the effectiveness of the proposed CSCLR, we first reproduce the AimCLR and conduct multiple experiments in all used datasets. Both reproduced results and the results reported in the paper are introduced for comparison, three streams (joint, bone and motion) are utilized for experiments. As shown in Table \ref{tab-2}, the proposed CSCLR achieves better performance than both AimCLR and AimCLR (Repro.), which validates the effectiveness of the CSCLR model. For single-stream results, CSCLR outperforms the AimCLR in all used stream, and especially achieves significant improvements in bone and motion stream. For the multi-stream fusion results, CSCLR also performs better than AimCLR in all used dataset.

\begin{table}[h]
	\centering
	\caption{\textrm{Ablation study results on NTU-60 dataset. The best results is in bold face.}}
	\label{tab-1}
	\renewcommand{\arraystretch}{1.2}
    \scalebox{0.9}{
	\begin{tabular}{ccccccc}
		\hline
		\multirow{2}{*}{Protocol}&\multirow{2}{*}{w/ Intra.} & \multirow{2}{*}{w/ Inter.} & \multirow{2}{*}{w/ PFT} & \multicolumn{2}{c}{NTU-60 (\%)} \\ \cline{5-6} 
		&&                                &                                     & xsub         & xview  \\ \hline
		\specialrule{0em}{1.2pt}{1.2pt}
		\multirow{3}{*}{\begin{tabular}[c]{@{}c@{}}Linear\\Eval.\end{tabular}}&
		\checkmark&         &                    &78.6           & 83.2       \\
		&\checkmark&      \checkmark&            &79.2           & 84.1              \\
		&\checkmark&      \checkmark&  \checkmark&\bfseries80.1           & \bfseries85.2              \\ \hline
		\specialrule{0em}{1.2pt}{1.2pt}
		\multirow{3}{*}{\begin{tabular}[c]{@{}c@{}}Finetuned\\Eval.\end{tabular}}&
		\checkmark&         &                    &85.4           & 91.2       \\
		&\checkmark&      \checkmark&            &86.2           & 91.9              \\
		&\checkmark&      \checkmark&  \checkmark&\bfseries87.0           & \bfseries 93.6              \\ \hline
	\end{tabular}
    }
\end{table}

\begin{figure}[t]
	\centering
	\includegraphics[width=7.71cm,height=6.5cm]{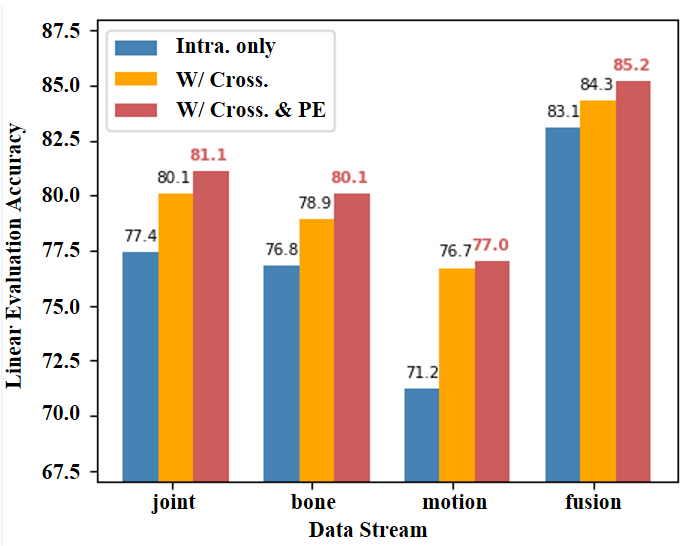}
	\caption{Linear evaluation results with different data stream on NTU-60-xview dataset.}
	\label{fig5}
\end{figure}

\textbf{The Effectiveness of Inter-stream Contrastive Learning.}
To further verify the proposed components in CSCLR, we conduct ablation studies on NTU-60 dataset and show the three-stream fusion results. As shown in Table \ref{tab-1}, when only utilizing the intra-stream contrastive learning (w/ Intra.), we can achieve the recognition accuracy of 78.4\% and 83.1\% on xsub and xview respectively. After adding the inter-stream contrastive learning (w/ Inter.), the performances are boosted, reaching 79.2\% and 84.1\%. This illustrates that the inter-stream alignment of hard positive pairs benefits the transfer accuracy of downstream recognition task. Similar to linear evaluation results, the performances of finetuned evaluation are also improved. The baseline accuracies (w/ Intra.) are 85.4\% and 91.2\%, then they are boosted up to 86.2\% and 91.9\% respectively. Besides, the specific improvements on different streams and the fusion result on NTU-60-xview dataset are shown in Figure \ref{fig5}. When using the combination of intra-stream and inter-stream contrastive learning, the linear evaluation accuracy of each stream is obviously boosted, the improvements on these streams are 2.7\%, 2.1\% and 5.5\% respectively. The general improvements on all used streams confirm the effectiveness of the proposed inter-stream contrastive learning method.

\textbf{The effectiveness of Positive Feature Transformation.}
From Table \ref{tab-1}, we can observe that when adopting positive feature transformation (w/ PFT), the linear evaluation accuracy on xsub and xview are further improved by 0.9\% and 1.1\%. A similar improvement can be observed when using the finetuned evaluation protocol, the accuracy are improved by 0.8\% and 1.7\%. Also, the specific improvements of PFT on each used skeleton data stream are shown in Figure \ref{fig5}. The improvements on joint, bone and motion stream are 1.2\%, 1.2\% and 0.3\% respectively. Due to that the original positive pairs in motion stream have already been dissimilar, the gains of PFT in this stream is slightly inferior to that in other streams. The improvement shows that the proposed PFT can enforces model to learn more robust feature for downstream tasks by generating harder positive samples with reasonable feature transformation. 

\begin{figure*}[t]
	\centering
	\includegraphics[width=16.0cm,height=8.5cm]{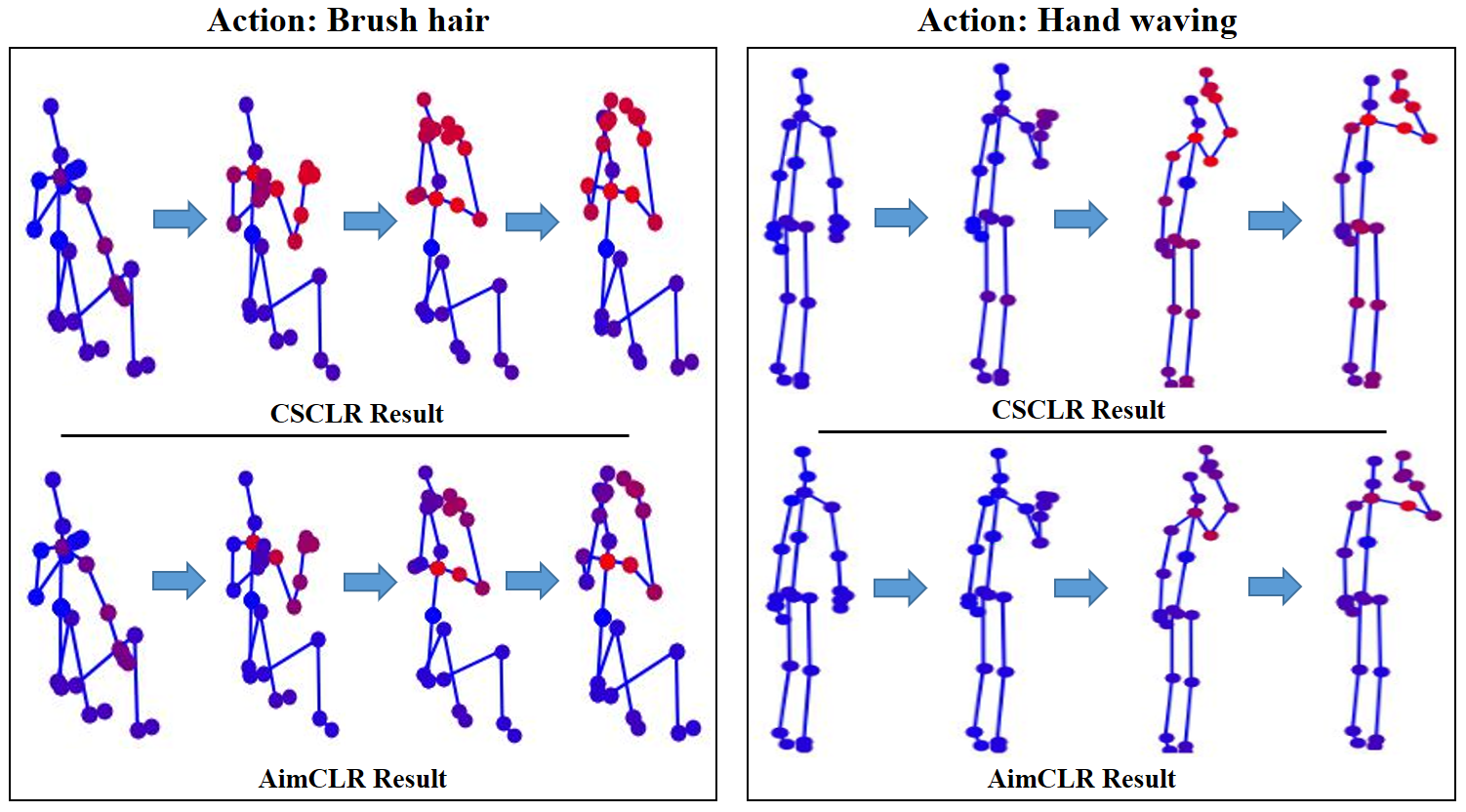}
	\caption{Skeleton Activation Map for sample actions \emph{Hand waving} and \emph{Brush hair}. The \textcolor{black}{red} joints are activated joints, and the \textcolor{black}{blue} joints are non-activated joints. (Best viewed in color.)}
	\label{fig7}
\end{figure*}
\textcolor{black}{\textbf{Skeleton Activation Map.} To show how CSCLR works, we utilize CAM (Class Activation Map) \cite{zhou2016learning} to calculate the activation map of skeleton sequence, and the activated joints in several frames are displayed in Figure \ref{fig7}. From this figure, we can observe that CSCLR is able to concentrate on the informative joints and thus achieve better performance. For example, the arms and hands are relatively informative for the action of brushing hair and hand waving. Compared with AimCLR, CSCLR pays higher attention to these informative body parts. The differences of skeleton activation map verify that CSCLR is conducive to learn effective features for downstream tasks.} 

\textcolor{black}{\textbf{Choice of streams in InterCLR.} To explore the impact of streams used in InterCLR, we further conduct experiments for comparion. We first set the method which only utilizes IntraCLR as a baseline for comparison, and then add InterCLR loss with different streams. The comparison follows linear evaluation protocol, and the detailed experiment results on NTU-60 are shown in Table \ref{tab-8}, in which ``2s'' and ``3s'' represent two-stream and three-stream respectively. From this table, we can observe that the accuracy performances are obviously boosted when adding 2s-InterCLR. When replacing 2s-InterCLR with 3s-InterCLR, the accuracy performances can be further improved, reaching 80.1\% and 85.2\% on xsub and xview respectively. The improvements again demonstrate the effectiveness of our proposed method.}
\begin{table}[ht]
	\centering
	\caption{\textrm{Linear evaluation results with different streams which are used in InterCLR.}}
	\label{tab-8}
	\renewcommand{\arraystretch}{1.2}
    \textrm{
	\begin{tabular}{cccccc}
		\toprule
		\multirow{2}{*}{Method} & \multicolumn{3}{c}{Stream in Inter.}& \multicolumn{2}{c}{NTU-60 (\%)}\\
        \cmidrule(r){2-4}
        \cmidrule(r){5-6}
		&joint&  bone &  motion   & xsub & xview  \\
        \midrule
        Base&-    &-         &-                    &78.6           &83.2       \\
        \midrule
		\multirow{3}{*}{\begin{tabular}[c]{@{}c@{}}With\\2s-Inter.\end{tabular}}&\checkmark&      \checkmark&            &79.7           &84.8               \\
        &\checkmark&      &            \checkmark&79.3           &84.5               \\
        &&      \checkmark&            \checkmark&79.0           &84.1               \\
        \midrule
		\begin{tabular}[c]{@{}c@{}}With\\3s-Inter.\end{tabular}&\checkmark&      \checkmark&  \checkmark&\bfseries80.1           & \bfseries85.2              \\
        \bottomrule
	\end{tabular}
    }
\end{table}

\textbf{Choice of Parameters in $F\left(\alpha, \mu\right)$.} As described in Section \ref{sec-3-4}, $\lambda \sim F\left(\alpha, \mu\right), F\left(\alpha, \mu\right) = Beta\left(\alpha, \alpha\right)\times \mu + 1$. To determine the hyper-parameters $\alpha$, $\mu$, we study how these parameters impact the performance of contrastive learning. As shown in Table \ref{tab-9}, we found that the best setting is $\alpha=2.0$, $\mu=1.0$. When $\alpha=0.5$, $\lambda$ tends to be close to the upper bound $\mu$ or lower bound $1+\mu$, the accuracy of recognition is invariant to the choice of $\mu$. When $\alpha=2.0$, $\lambda$ is sampled to be $1+\mu/2$ with high probability, better accuracy are obtained. Therefore, we choose $\alpha=2.0$, $\mu=1.0$ as default setting.
\begin{table}[h]
	\centering
	\caption{\textrm{Results of CSCLR-joint with various $\alpha$, $\mu$ on NTU-60 dataset. The best results is in bold face.}}
    \label{tab-9}
	\renewcommand{\arraystretch}{1.2}
	\begin{tabular}{cccc}
		\hline
		\multirow{2}{*}{$\alpha$} & \multirow{2}{*}{$\mu$} & \multicolumn{2}{c}{\textrm{NTU-60-joint (\%)}} \\ \cline{3-4} 
		&&                                                         \textrm{xsub}         & \textrm{xview}  \\ \hline
		\specialrule{0em}{1.2pt}{1.2pt}
		
		\multirow{3}{*}{\textrm{0.5}}    &\textrm{1.0}  & \textrm{75.0} & \textrm{80.9}    \\
		&\textrm{1.2}            & \textrm{74.9} & \textrm{80.8}                        \\
		&\textrm{1.5}  & \textrm{74.9} & \textrm{80.9}                      \\ \hline
		\specialrule{0em}{1.2pt}{1.2pt}
		\multirow{3}{*}{\textrm{2.0}}  &\textrm{1.0}  & \bfseries\textrm{75.7}  & \bfseries\textrm{81.3}                 \\
		&\textrm{1.2}            & \textrm{75.0}  & \textrm{81.1}                     \\
		&\textrm{1.5}            & \textrm{75.2}  & \textrm{81.0}                    \\ \hline
	\end{tabular}
\end{table}

\section{Conclusion}\label{sec-5}
In this paper, we advocate the significant effect of inter-stream contrast pairs, and therefore propose a Cross-Stream Contrastive Learning framework for skeleton-based action Representation learning (CSCLR). Except for utilizing easy contrast pairs within a single data stream, CSCLR additionally introduces inter-stream contrast pairs as hard samples to formulate a better representation learning. Besides, to conduct in-depth study on positive samples, we propose a Positive Feature Transformation (PFT) strategy which adopts feature-level manipulation to increase the variance of positive pairs. \textcolor{black}{Extensive experiment results on NTU-RGB+D 60, NTU-RGB+D 120 and PKU-MMD demonstrate that CSCLR achieves superior classification accuracy on linear evaluation, finetuned evaluation and semi-supervised evaluation.} The ablation study demonstrates that both the inter-stream hard contrast pairs and the PFT strategy can promote more robust representation learning. In the future work, we will continue to explore the design of contrast pairs specified for skeleton-based tasks. \textcolor{black}{The unified data augmentations are used for different skeleton data streams, which ignore the discrepancy between them. Thus, with the customized data augmentations for each data stream, the performance of downstream tasks would be further improved. } 

{\small
\bibliographystyle{ieee_fullname}
\bibliography{egbib}

\begin{thebibliography}{10}\itemsep=-1pt

\bibitem{afham2022crosspoint}
Mohamed Afham, Isuru Dissanayake, Dinithi Dissanayake, Amaya Dharmasiri,
  Kanchana Thilakarathna, and Ranga Rodrigo.
\newblock Crosspoint: Self-supervised cross-modal contrastive learning for 3d
  point cloud understanding.
\newblock In {\em Proceedings of the IEEE/CVF Conference on Computer Vision and
  Pattern Recognition}, pages 9902--9912, 2022.

\bibitem{arandjelovic2017look}
Relja Arandjelovic and Andrew Zisserman.
\newblock Look, listen and learn.
\newblock In {\em Proceedings of the IEEE/CVF International Conference on
  Computer Vision}, pages 609--617, 2017.

\bibitem{chen2020simple}
Ting Chen, Simon Kornblith, Mohammad Norouzi, and Geoffrey Hinton.
\newblock A simple framework for contrastive learning of visual
  representations.
\newblock {\em arXiv preprint arXiv:2002.05709}, 2020.

\bibitem{chen2020big}
Ting Chen, Simon Kornblith, Kevin Swersky, Mohammad Norouzi, and Geoffrey
  Hinton.
\newblock Big self-supervised models are strong semi-supervised learners.
\newblock {\em arXiv preprint arXiv:2006.10029}, 2020.

\bibitem{chen2020improved}
Xinlei Chen, Haoqi Fan, Ross Girshick, and Kaiming He.
\newblock Improved baselines with momentum contrastive learning.
\newblock {\em arXiv preprint arXiv:2003.04297}, 2020.

\bibitem{chen2021channel}
Yuxin Chen, Ziqi Zhang, Chunfeng Yuan, Bing Li, Ying Deng, and Weiming Hu.
\newblock Channel-wise topology refinement graph convolution for skeleton-based
  action recognition.
\newblock In {\em Proceedings of the IEEE/CVF Conference on Computer Vision and
  Pattern Recognition}, pages 13359--13368, 2021.

\bibitem{cheng2021motion}
Yi-Bin Cheng, Xipeng Chen, Dongyu Zhang, and Liang Lin.
\newblock Motion-transformer: self-supervised pre-training for skeleton-based
  action recognition.
\newblock In {\em Proceedings of the 2nd ACM International Conference on
  Multimedia in Asia}, pages 1--6, 2021.

\bibitem{chi2022infogcn}
Hyung-gun Chi, Myoung~Hoon Ha, Seunggeun Chi, Sang~Wan Lee, Qixing Huang, and
  Karthik Ramani.
\newblock Infogcn: Representation learning for human skeleton-based action
  recognition.
\newblock In {\em Proceedings of the IEEE/CVF Conference on Computer Vision and
  Pattern Recognition}, pages 20186--20196, 2022.

\bibitem{desai2021virtex}
Karan Desai and Justin Johnson.
\newblock Virtex: Learning visual representations from textual annotations.
\newblock In {\em Proceedings of the IEEE/CVF Conference on Computer Vision and
  Pattern Recognition}, pages 11162--11173, 2021.

\bibitem{ding2021kfc}
Xinpeng Ding, Nannan Wang, Xinbo Gao, Jie Li, Xiaoyu Wang, and Tongliang Liu.
\newblock Kfc: An efficient framework for semi-supervised temporal action
  localization.
\newblock {\em IEEE Trans. Image Proc.}, 30:6869--6878, 2021.

\bibitem{du2015hierarchical}
Yong Du, Wei Wang, and Liang Wang.
\newblock Hierarchical recurrent neural network for skeleton based action
  recognition.
\newblock In {\em Proceedings of the IEEE/CVF Conference on Computer Vision and
  Pattern Recognition}, pages 1110--1118, 2015.

\bibitem{freire2022inflated}
David Freire-Obreg{\'o}n, Paola Barra, Modesto Castrill{\'o}n-Santana, and
  Maria~De Marsico.
\newblock Inflated 3d convnet context analysis for violence detection.
\newblock {\em Machine Vision and Applications}, 33:1--13, 2022.

\bibitem{gao2021efficient}
Xuehao Gao, Yang Yang, Yimeng Zhang, Maosen Li, Jin-Gang Yu, and Shaoyi Du.
\newblock Efficient spatio-temporal contrastive learning for skeleton-based 3d
  action recognition.
\newblock {\em IEEE Trans. Multi.}, 2021.
\newblock early access.

\bibitem{guo2022contrastive}
Tianyu Guo, Hong Liu, Zhan Chen, Mengyuan Liu, Tao Wang, and Runwei Ding.
\newblock Contrastive learning from extremely augmented skeleton sequences for
  self-supervised action recognition.
\newblock In {\em Proceedings of the AAAI Conference on Artificial
  Intelligence}, pages 762--770, 2022.

\bibitem{he2020momentum}
Kaiming He, Haoqi Fan, Yuxin Wu, Saining Xie, and Ross Girshick.
\newblock Momentum contrast for unsupervised visual representation learning.
\newblock In {\em Proceedings of the IEEE/CVF Conference on Computer Vision and
  Pattern Recognition}, pages 9729--9738, 2020.

\bibitem{hjelm2018learning}
R~Devon Hjelm, Alex Fedorov, Samuel Lavoie-Marchildon, Karan Grewal, Phil
  Bachman, Adam Trischler, and Yoshua Bengio.
\newblock Learning deep representations by mutual information estimation and
  maximization.
\newblock {\em arXiv preprint arXiv:1808.06670}, 2018.

\bibitem{hochreiter2001gradient}
Sepp Hochreiter, Yoshua Bengio, Paolo Frasconi, J{\"u}rgen Schmidhuber, et~al.
\newblock {\em Gradient flow in recurrent nets: the difficulty of learning
  long-term dependencies}.
\newblock A field guide to dynamical recurrent neural networks. IEEE Press,
  2001.

\bibitem{huang2021self}
Jing Huang, Yan Huang, Qicong Wang, Wenming Yang, and Hongying Meng.
\newblock Self-supervised representation learning for videos by segmenting via
  sampling rate order prediction.
\newblock {\em IEEE Trans. Circ. Syst. Video Tech.}, 32(6):3475--3489, 2021.

\bibitem{ke2017new}
Qiuhong Ke, Mohammed Bennamoun, Senjian An, Ferdous Sohel, and Farid Boussaid.
\newblock A new representation of skeleton sequences for 3d action recognition.
\newblock In {\em Proceedings of the IEEE/CVF Conference on Computer Vision and
  Pattern Recognition}, pages 3288--3297, 2017.

\bibitem{li2017skeleton}
Chao Li, Qiaoyong Zhong, Di Xie, and Shiliang Pu.
\newblock Skeleton-based action recognition with convolutional neural networks.
\newblock In {\em IEEE International Conference on Multimedia \& Expo
  Workshops}, pages 597--600, 2017.

\bibitem{li20213d}
Linguo Li, Minsi Wang, Bingbing Ni, Hang Wang, Jiancheng Yang, and Wenjun
  Zhang.
\newblock 3d human action representation learning via cross-view consistency
  pursuit.
\newblock In {\em Proceedings of the IEEE/CVF Conference on Computer Vision and
  Pattern Recognition}, pages 4741--4750, 2021.

\bibitem{li2018independently}
Shuai Li, Wanqing Li, Chris Cook, Ce Zhu, and Yanbo Gao.
\newblock Independently recurrent neural network (indrnn): Building a longer
  and deeper rnn.
\newblock In {\em Proceedings of the IEEE/CVF Conference on Computer Vision and
  Pattern Recognition}, pages 5457--5466, 2018.

\bibitem{lin2020ms2l}
Lilang Lin, Sijie Song, Wenhan Yang, and Jiaying Liu.
\newblock Ms2l: Multi-task self-supervised learning for skeleton based action
  recognition.
\newblock In {\em Proceedings of the 28th ACM International Conference on
  Multimedia}, pages 2490--2498, 2020.

\bibitem{liu2019ntu}
Jun Liu, Amir Shahroudy, Mauricio Perez, Gang Wang, Ling-Yu Duan, and Alex~C
  Kot.
\newblock Ntu rgb+ d 120: A large-scale benchmark for 3d human activity
  understanding.
\newblock {\em IEEE Trans. Pat. Anal. Machi. Intel.}, 42(10):2684--2701, 2019.

\bibitem{liu2017skeleton}
Jun Liu, Gang Wang, Ling-Yu Duan, Kamila Abdiyeva, and Alex~C Kot.
\newblock Skeleton-based human action recognition with global context-aware
  attention lstm networks.
\newblock {\em IEEE Trans. Image Proc.}, 27(4):1586--1599, 2017.

\bibitem{tgl9713748}
Yang Liu, Keze Wang, Lingbo Liu, Haoyuan Lan, and Liang Lin.
\newblock Tcgl: Temporal contrastive graph for self-supervised video
  representation learning.
\newblock {\em IEEE Trans. Image Proc.}, 31:1978--1993, 2022.

\bibitem{liu2020disentangling}
Ziyu Liu, Hongwen Zhang, Zhenghao Chen, Zhiyong Wang, and Wanli Ouyang.
\newblock Disentangling and unifying graph convolutions for skeleton-based
  action recognition.
\newblock In {\em Proceedings of the IEEE/CVF Conference on Computer Vision and
  Pattern Recognition}, pages 143--152, 2020.

\bibitem{misra2020self}
Ishan Misra and Laurens van~der Maaten.
\newblock Self-supervised learning of pretext-invariant representations.
\newblock In {\em Proceedings of the IEEE/CVF Conference on Computer Vision and
  Pattern Recognition}, pages 6707--6717, 2020.

\bibitem{morgado2021audio}
Pedro Morgado, Nuno Vasconcelos, and Ishan Misra.
\newblock Audio-visual instance discrimination with cross-modal agreement.
\newblock In {\em Proceedings of the IEEE/CVF Conference on Computer Vision and
  Pattern Recognition}, pages 12475--12486, 2021.

\bibitem{nie2020unsupervised}
Qiang Nie, Ziwei Liu, and Yunhui Liu.
\newblock Unsupervised 3d human pose representation with viewpoint and pose
  disentanglement.
\newblock In {\em Proceedings of the European Conference on Computer Vision},
  pages 102--118, 2020.

\bibitem{oord2018representation}
Aaron van~den Oord, Yazhe Li, and Oriol Vinyals.
\newblock Representation learning with contrastive predictive coding.
\newblock {\em arXiv preprint arXiv:1807.03748}, 2018.

\bibitem{owens2018audio}
Andrew Owens and Alexei~A Efros.
\newblock Audio-visual scene analysis with self-supervised multisensory
  features.
\newblock In {\em Proceedings of the European Conference on Computer Vision},
  pages 631--648, 2018.

\bibitem{pan2021videomoco}
Tian Pan, Yibing Song, Tianyu Yang, Wenhao Jiang, and Wei Liu.
\newblock Videomoco: Contrastive video representation learning with temporally
  adversarial examples.
\newblock In {\em Proceedings of the IEEE/CVF Conference on Computer Vision and
  Pattern Recognition}, pages 11205--11214, 2021.

\bibitem{paszke2017automatic}
Adam Paszke, Sam Gross, Soumith Chintala, Gregory Chanan, Edward Yang, Zachary
  DeVito, Zeming Lin, Alban Desmaison, Luca Antiga, and Adam Lerer.
\newblock Automatic differentiation in pytorch.
\newblock In {\em Advances in Neural Information Processing Systems}, pages
  8026--8037, 2017.

\bibitem{presti20163d}
Liliana~Lo Presti and Marco La~Cascia.
\newblock 3d skeleton-based human action classification: A survey.
\newblock {\em Pattern Recognition}, 53:130--147, 2016.

\bibitem{radford2021learning}
Alec Radford, Jong~Wook Kim, Chris Hallacy, Aditya Ramesh, Gabriel Goh,
  Sandhini Agarwal, Girish Sastry, Amanda Askell, Pamela Mishkin, Jack Clark,
  et~al.
\newblock Learning transferable visual models from natural language
  supervision.
\newblock {\em arXiv preprint arXiv:2103.00020}, 2021.

\bibitem{rao2021augmented}
Haocong Rao, Shihao Xu, Xiping Hu, Jun Cheng, and Bin Hu.
\newblock Augmented skeleton based contrastive action learning with momentum
  lstm for unsupervised action recognition.
\newblock {\em Information Sciences}, 569:90--109, 2021.

\bibitem{sariyildiz2020learning}
Mert~Bulent Sariyildiz, Julien Perez, and Diane Larlus.
\newblock Learning visual representations with caption annotations.
\newblock In {\em Proceedings of the European Conference on Computer Vision},
  pages 153--170. Springer, 2020.

\bibitem{shi2019skeleton}
Lei Shi, Yifan Zhang, Jian Cheng, and Hanqing Lu.
\newblock Skeleton-based action recognition with directed graph neural
  networks.
\newblock In {\em Proceedings of the IEEE/CVF Conference on Computer Vision and
  Pattern Recognition}, pages 7912--7921, 2019.

\bibitem{shi2019two}
Lei Shi, Yifan Zhang, Jian Cheng, and Hanqing Lu.
\newblock Two-stream adaptive graph convolutional networks for skeleton-based
  action recognition.
\newblock In {\em Proceedings of the IEEE/CVF Conference on Computer Vision and
  Pattern Recognition}, pages 12026--12035, 2019.

\bibitem{song2017end}
Sijie Song, Cuiling Lan, Junliang Xing, Wenjun Zeng, and Jiaying Liu.
\newblock An end-to-end spatio-temporal attention model for human action
  recognition from skeleton data.
\newblock In {\em Proceedings of the AAAI Conference on Artificial
  Intelligence}, pages 4263--4270, 2017.

\bibitem{su2020predict}
Kun Su, Xiulong Liu, and Eli Shlizerman.
\newblock Predict \& cluster: Unsupervised skeleton based action recognition.
\newblock In {\em Proceedings of the IEEE/CVF Conference on Computer Vision and
  Pattern Recognition}, pages 9631--9640, 2020.

\bibitem{sun2021multi}
Ning Sun, Ling Leng, Jixin Liu, and Guang Han.
\newblock Multi-stream slowfast graph convolutional networks for skeleton-based
  action recognition.
\newblock {\em Image Vis. Comput.}, 109:1014--1030, 2021.

\bibitem{thoker2021skeleton}
Fida~Mohammad Thoker, Hazel Doughty, and Cees~GM Snoek.
\newblock Skeleton-contrastive 3d action representation learning.
\newblock In {\em Proceedings of the 29th ACM International Conference on
  Multimedia}, pages 1655--1663, 2021.

\bibitem{tian2020makes}
Yonglong Tian, Chen Sun, Ben Poole, Dilip Krishnan, Cordelia Schmid, and
  Phillip Isola.
\newblock What makes for good views for contrastive learning?
\newblock {\em arXiv preprint arXiv:2005.10243}, 2020.

\bibitem{van2008visualizing}
Laurens Van~der Maaten and Geoffrey Hinton.
\newblock Visualizing data using t-sne.
\newblock {\em Jour. Mach. Learn. Res.}, 9(11):2579--2605, 2008.

\bibitem{vemulapalli2014human}
Raviteja Vemulapalli, Felipe Arrate, and Rama Chellappa.
\newblock Human action recognition by representing 3d skeletons as points in a
  lie group.
\newblock In {\em Proceedings of the IEEE/CVF Conference on Computer Vision and
  Pattern Recognition}, pages 588--595, 2014.

\bibitem{vemulapalli2016rolling}
Raviteja Vemulapalli and Rama Chellapa.
\newblock Rolling rotations for recognizing human actions from 3d skeletal
  data.
\newblock In {\em Proceedings of the IEEE/CVF Conference on Computer Vision and
  Pattern Recognition}, pages 4471--4479, 2016.

\bibitem{verma2018manifold}
Vikas Verma, Alex Lamb, Christopher Beckham, Amir Najafi, Ioannis Mitliagkas,
  Aaron Courville, David Lopez-Paz, and Yoshua Bengio.
\newblock Manifold mixup: Better representations by interpolating hidden
  states.
\newblock {\em arXiv preprint arXiv:1806.05236}, 2018.

\bibitem{Wang_2022_CVPR}
Haoqing Wang, Xun Guo, Zhi-Hong Deng, and Yan Lu.
\newblock Rethinking minimal sufficient representation in contrastive learning.
\newblock In {\em Proceedings of the IEEE/CVF Conference on Computer Vision and
  Pattern Recognition}, pages 16041--16050, June 2022.

\bibitem{wang2012mining}
Jiang Wang, Zicheng Liu, Ying Wu, and Junsong Yuan.
\newblock Mining actionlet ensemble for action recognition with depth cameras.
\newblock In {\em Proceedings of the IEEE/CVF Conference on Computer Vision and
  Pattern Recognition}, pages 1290--1297, 2012.

\bibitem{yan2018spatial}
Sijie Yan, Yuanjun Xiong, and Dahua Lin.
\newblock Spatial temporal graph convolutional networks for skeleton-based
  action recognition.
\newblock In {\em Proceedings of the AAAI Conference on Artificial
  Intelligence}, pages 7444--7452, 2018.

\bibitem{yang2021skeleton}
Siyuan Yang, Jun Liu, Shijian Lu, Meng~Hwa Er, and Alex~C Kot.
\newblock Skeleton cloud colorization for unsupervised 3d action representation
  learning.
\newblock In {\em Proceedings of the IEEE/CVF International Conference on
  Computer Vision}, pages 13423--13433, 2021.

\bibitem{9841515}
Yang Yang, Guangjun Liu, and Xuehao Gao.
\newblock Motion guided attention learning for self-supervised 3d human action
  recognition.
\newblock {\em IEEE Trans. Circ. Syst. Video Tech.}, 32(12):8623--8634, 2022.

\bibitem{zhang2017mixup}
Hongyi Zhang, Moustapha Cisse, Yann~N Dauphin, and David Lopez-Paz.
\newblock Mixup: Beyond empirical risk minimization.
\newblock {\em arXiv preprint arXiv:1710.09412}, 2017.

\bibitem{zheng2018unsupervised}
Nenggan Zheng, Jun Wen, Risheng Liu, Liangqu Long, Jianhua Dai, and Zhefeng
  Gong.
\newblock Unsupervised representation learning with long-term dynamics for
  skeleton based action recognition.
\newblock In {\em Proceedings of the AAAI Conference on Artificial
  Intelligence}, pages 2644--2651, 2018.

\bibitem{zhou2016learning}
Bolei Zhou, Aditya Khosla, Agata Lapedriza, Aude Oliva, and Antonio Torralba.
\newblock Learning deep features for discriminative localization.
\newblock In {\em Proceedings of the IEEE/CVF Conference on Computer Vision and
  Pattern Recognition}, pages 2921--2929, 2016.

\bibitem{zhou2021domain}
Kaiyang Zhou, Yongxin Yang, Yu Qiao, and Tao Xiang.
\newblock Domain generalization with mixstyle.
\newblock {\em arXiv preprint arXiv:2104.02008}, 2021.

\bibitem{zhu2021improving}
Rui Zhu, Bingchen Zhao, Jingen Liu, Zhenglong Sun, and Chang~Wen Chen.
\newblock Improving contrastive learning by visualizing feature transformation.
\newblock In {\em Proceedings of the IEEE/CVF Conference on Computer Vision and
  Pattern Recognition}, pages 10306--10315, 2021.

\end{thebibliography}
}

\end{document}